\documentclass[sigconf,nonacm]{acmart}
\input{preface.tex}

\AtBeginDocument{%
  \providecommand\BibTeX{{%
    \normalfont B\kern-0.5em{\scshape i\kern-0.25em b}\kern-0.8em\TeX}}}

\setcopyright{acmcopyright}
\copyrightyear{2020}
\acmYear{2020}
\acmDOI{10.1145/1122445.1122456}

\acmConference[CIKM'21]{SIGKDD}{November 1--5, 2021}{}
\acmPrice{15.00}
\acmISBN{978-1-4503-XXXX-X/18/06}

\begin{document}

\title{
    Dyport: Dynamic Importance-based Hypothesis Generation Benchmarking Technique
}

%Following KDD conference tradition, reviews are not double-blind, and author names and affiliations should be listed.

\author{Ilya Tyagin}
\orcid{0000-0002-0249-1072}
\affiliation{%
  \institution{Center for Bioinformatics and Computational Biology \\University of Delaware\\Newark, DE}
  %\streetaddress{821 McMillan Rd.}
  %\city{Clemson}
  %\state{SC}
  %\postcode{29630}
  \country{USA}
  }
\email{tyagin@udel.edu}

\author{Ilya Safro}
\orcid{0000-0001-6284-7408}
\affiliation{%
  \institution{Computer and Information Sciences\\University of Delaware\\Newark, DE}
  %\streetaddress{821 McMillan Rd.}
  %\city{Clemson}
  %\state{SC}
  %\postcode{29630}
  \country{USA}
  }
\email{isafro@udel.edu}

\begin{abstract}
    This paper presents a novel benchmarking framework \sysname for evaluating biomedical hypothesis generation systems. Utilizing curated datasets, our approach tests these systems under realistic conditions, enhancing the relevance of our evaluations. We integrate knowledge from the curated databases into a dynamic graph, accompanied by a method to quantify discovery importance. This not only assesses hypothesis accuracy but also their potential impact in biomedical research which significantly extends  traditional link prediction benchmarks. Applicability of our benchmarking process is demonstrated on several link prediction systems applied on biomedical semantic knowledge graphs. Being flexible, our benchmarking system is designed for broad application in hypothesis generation quality verification, aiming to expand the scope of scientific discovery within the biomedical research community.\\
    \textbf{Availability and implementation:} Dyport framework is fully open-source. All code and datasets are available at: \url{https://github.com/IlyaTyagin/Dyport}
\end{abstract}

\maketitle

\section{Introduction}
Automated hypothesis generation (HG, also known as Literature Based Discovery, LBD) has gone a long way since its establishment in 1986, when Swanson introduced the concept of "Undiscovered Public Knowledge"~\cite{swanson1986undiscovered}. It pertains to the idea that within the public domain, there is a significant abundance of information, allowing for the uncovering of implicit connections among various pieces of information.  There are many systems developed throughout the years, which incorporate different reasoning methods: from  concept co-occurrence in scientific literature \cite{swanson2006ranking} to the advanced deep learning-based algorithms and generative models (such as BioGPT~\cite{luo2022biogpt} and CBAG \cite{sybrandt2021cbag}). Examples include but are not limited to probabilistic topic modeling over relevant papers~\cite{sybrandt2017}, semantic inference \cite{sedler2019semnet}, association rule discovery~\cite{hristovski2005using}, latent semantic indexing \cite{gordon1998using}, semantic knowledge network completion~\cite{sybrandt2020agatha} to mention just a few. The common thread running through these lines of research is that they are all meant to fill in the gaps between pieces of existing knowledge.

The evaluation of HG is still one of the major problems of these systems, especially when it comes to fully automated large-scale general purpose systems (such as IBM Watson Drug Discovery~\cite{chen2016ibm}, AGATHA \cite{sybrandt2020agatha} or BioGPT \cite{luo2022biogpt}). For these,
a massive assessment (that is normal in the machine learning and general AI domains) performed manually by the domain experts is usually not feasible and other methods are required.%, especially in the earlier development stages.  

One traditional evaluation approach is to make a system ``rediscover'' some of the landmark findings, similar to was done in numerous works replicating well-known connections, such as: \textit{Fish Oil $\leftrightarrow$ Raynaud's Syndrome}~\cite{8215526}, \emph{Migraine $\leftrightarrow$ Magnesium} ~\cite{8215526} or \emph{Alzheimer $\leftrightarrow$ Estrogen}~\cite{CAMERON2015141}. This technique is frequently used even in a majority of the recently published papers, despite of its obvious drawbacks, such as very limited number of validation samples and their general obsolesce (some of these connections are over 30 years old) not to mention that in some of these works, the training set is not carefully chosen to include only the information published prior the discovery of interest which turns the HG goal into the information retrieval task. 
    
Another commonly used technique is based on the time-slicing, when a system is trained on a subset of data prior to a specified cut-off date and then evaluated on the data from the future. This method addresses the weaknesses of previous approach and can be automated, but it does not immediately answer the question of how significant or impactful the connections are. The lack of this information may lead to deceiving results: many connections, even recently published, are trivial (especially if they are found by the text mining methods) and do not advance the scientific field in a meaningful way.

A related area that faces similar evaluation challenges is Information Extraction (IE), a field crucial to enabling effective HG by identifying and categorizing relevant information in publicly available data sources. Within the realm of biomedical and life sciences IE, there are more targeted, small-scale evaluation protocols such as the BioCreative competitions~\cite{miranda2021overview}, where the domain experts provide curated training and test datasets, which allows participants to refine and assess their systems within a controlled environment. While such targeted evaluations as conducted in BioCreative are both crucial and insightful, they inherently lack the scope and scale needed for the evaluation of expansive HG systems. 

The aforementioned issues emphasize the critical need for research into effective, scalable evaluation methods in automated hypothesis generation. Our primary interest is in establishing an effective and sustainable benchmark for large-scale, general-purpose automated hypothesis generation systems within the biomedical domain. We seek to identify substantial, non-trivial insights, prioritizing them over mere data volume and ensuring scalability with respect to ever-expanding biocurated knowledge databases. We emphasize the significance of implementing sustainable evaluation strategies, relying on constantly updated datasets reflecting the latest research. Lastly, our efforts are targeted towards distinguishing between hypotheses with significant impact and those with lesser relevance, thus moving beyond trivial generation of hypotheses to ensuring their meaningful contribution to scientific discovery.\\
    
\noindent {\bf Our contribution}:
\begin{itemize}
\item We propose a high quality benchmark dataset \sysname for hypothesis prediction systems evaluation. 

It incorporates information extracted from a number of biocurated databases. We normalize all concepts to the unified format for seamless integration and each connection is supplied with rich metadata, including timestamp information to enable time-slicing. 

\item We introduce an evaluation method for the impact of connections in time-slicing paradigm. It will allow to benchmark HG systems more thoroughly and extensively by assigning an importance weight to every connection over the time. This weight represents the overall impact a connection makes on future discovery.

\item We demonstrate the computational results of several prediction algorithms using the proposed benchmark and discuss their performance and quality.
\end{itemize} 

We propose to use our benchmark to evaluate the quality of HG systems. The benchmark will be updated on at least a yearly basis. Its structure facilitates relatively effortless expansion by other researchers.
\section{Background and Related Work}
%\noindent\textbf{Benchmarking Hypothesis Generation Systems.}
Unfortunately, the evaluation in the hypothesis generation field is often coupled with the systems to evaluate and currently not universally standardized. If one would like to compare the performance of two or more systems, they need to understand their training protocol to instantiate models from scratch and then test them on the same data they used in their experiment.

This problem is well known and there are attempts to provide a universal way to evaluate such systems. For example, OpenBioLink~\cite{10.1093/bioinformatics/btaa274} is designed as a software package for evaluation of link prediction models. It supports time-slicing and contains millions of edges with different quality settings. The authors describe it as ``highly challenging'' dataset that does not include  "trivially predictable" connections, but they do not provide a quantification of 
difficulty nor range the edges accordingly.

Another attempt to set up a large-scale validation of HG systems was performed in our earlier work~\cite{sybrandt2018a}. The proposed methodology is based on the semantic triples extracted from SemMedDB~\cite{KilicogluSFRR12} database and setting up a cut date for training and testing. Triples are converted to pairs by removing the ``verb'' part from each \textit{(subject-verb-object)} triple. For the test data, a list of "highly cited" pairs is identified, which is based on the citation counts from SemMedDB, MEDLINE and Semantic Scholar. Only connections occurring in papers published after the cut date and cited over 100 times are considered. It is worth mentioning that this approach is prone to noise (due to SemMedDB text mining methods) and also skewed towards the discoveries published closer to the cut-date, since the citations accumulate over time.

Currently, Knowledge Graph Embeddings (KGE) are becoming increasingly popular and the hypothesis generation problem can be formulated in terms of the link prediction in knowledge graphs. Knowledge Graphs often evaluate the likelihood of a particular connection with the scoring function of choice. For example, TransE~\cite{bordes2013translating} evaluates each sample with the following equation:
\begin{equation*}
        s(h, r, t) = \| \mathbf{h} + \mathbf{r} - \mathbf{t} \|,
\end{equation*}
where $h$ is the embedding vector of a head entity, $r$ is the embedding vector of relation, $t$ is the embedding vector of a tail entity and $||\cdot||$ denotes the L1 or L2 norm. 
    
These days KGE-based models are of interest to the broad scientific community, including researchers in the drug discovery field. Recently they carefully investigated the factors affecting the performance of KGE models~\cite{bonner2022understanding} and reviewed biomedical databases related to drug discovery~\cite{bonner2022review}. These publications, however, do not focus on any temporal information nor attempt to describe the extracted concept associations quantitatively. We also aim to fill in this currently existing gap in our current work.

\section{Incorporated Technologies}
\label{sec:incorp_technologies}
To construct the benchmark, we propose a multi-step pipeline, which requires several key technologies to be used. For the text mining part, we use SemRep~\cite{RINDFLESCH2003462} and gensim~\cite{rehurek2011gensim} implementation of word2vec algorithm. For further stages involving graph learning, we utilize Pytorch Geometric framework and Captum explainability library. 

\noindent\textbf{UMLS (Unified Medical Language System)}~\cite{bodenreider2004unified} is one of the fundamental technologies provided by NLM, which consolidates and disseminates essential terminology, taxonomies, and coding norms, along with related materials, such as definitions and semantic types. UMLS is used in the proposed work as a system of concept unique identifiers (CUI) bringing together terms from different vocabularies.

\noindent\textbf{SemRep~\cite{RINDFLESCH2003462}} is an NLM-developed software, performing extraction of semantic predicates from biomedical texts. It also has the named entity recognition (NER) capabilities (based on MetaMap~\cite{aronson2001effective} backend) and automatically performs entity normalization based on the context. 
    
\noindent\textbf{Word2Vec}~\cite{mikolov2013distributed} is an approach for creating efficient word embeddings. It was proposed in 2013 and is proven to be an excellent technique for generating static (context-independent) latent word representations. The implementation used in this work is based on gensim~\cite{rehurek2011gensim} library.
    
\noindent\textbf{Pytorch Geometric (PyG)~\cite{Fey/Lenssen/2019}} library is built on top of Pytorch framework focusing on graph geometric learning. It implements a variety of algorithms from published research papers, supports arbitrary-scaled graphs and is well integrated into Pytorch ecosystem. We use PyG to train a graph neural network (GNN) for link prediction problem, which we explain in more detail in methods section.

\noindent\textbf{Captum~\cite{kokhlikyan2020captum}} package is an extension of Pytorch enabling the explainability of many ML models. It contains attribution methods, such as saliency maps, integrated gradients, Shapley value sampling and others. Captum is supported by PyG library and used in this work to calculate attributions of the proposed GNN.
    
\section{Incorporated Data Sources}
We review and include a variety of biomedical databases, containing curated connections between different kinds of entities. 
    
\noindent\textbf{KEGG (Kyoto Encyclopedia of Genes and Genomes)~\cite{10.1093/nar/gkv1070}} is a collection of resources for understanding principles of work of biological systems (such as cells, organisms or ecosystems) and offering a wide variety of entry points. One of the main components of KEGG is a set of pathway maps, representing molecular interactions as network diagrams. 
    
\noindent\textbf{CTD (The Comparative Toxicogenomics Database)~\cite{davis_wiegers_johnson_sciaky_wiegers_mattingly_2022}} is a publicly available database focused on collecting the information about environmental exposures effects on human health. 
    
\noindent\textbf{DisGenNET~\cite{10.1093/nar/gkz1021}} is a discovery platform covering genes and variants and their connections to human diseases. It integrates data from a list of publicly available databases and repositories and scientific literature. 
    
\noindent\textbf{GWAS (Genome-Wide Association Studies)~\cite{10.1093/nar/gkac1010}} is a catalog of human genome-wide association studies, developed by EMBL-EBI and NHGRI. Its aim is to identify and systematize associations of genotypes with phenotypes across human genome. 
    
\noindent\textbf{STRING~\cite{10.1093/nar/gkaa1074}} is a database aiming to integrate known and predicted protein associations, both physical and functional. It utilizes a network-centric approach and assigns a confidence score for all interactions in the network based on the evidence coming from different sources: text mining, computational predictions and biocurated databases.
    
\noindent\textbf{DrugCentral~\cite{10.1093/nar/gkw993}} is an online drug information resource aggregating information about active ingredients, indications, pharmacologic action and other related data with respect to FDA, EMA and PMDA-approved drugs. 
    
\noindent\textbf{Mentha~\cite{calderone2013mentha}} is an evidence-based protein interaction browser (and corresponding database), which takes advantage of International Molecular Exchange (IMEx) consortium. The interactions are curated by experts in compliance with IMEx policies enabling regular weekly updates. Compared to STRING, Mentha is focused on precision over comprehensiveness and excludes any computationally predicted records. 
    
\noindent\textbf{RxNav~\cite{zeng2006rxnav}} is a web-service providing an integrated view on drug information. It contains the information from NLM drug terminology RxNorm, drug classes RxClass and drug-drug interactions collected from ONCHigh and DrugBank sources. 

\noindent\textbf{Semantic Scholar~\cite{fricke2018semantic}} is a search engine and research tool for scientific papers developed by the Allen Institute for Artificial Intelligence (AI2). It provides rich metadata about publications which enables us to use Semantic Scholar data for network-based citation analysis.
\section{Methods}

\subsection{Glossary}
\begin{itemize}
    \item $c_i$ - concept in some arbitrary vocabulary;
    \item $m(\cdot)$ - function that maps a concept $c_i$ to the subset of corresponding  UMLS CUI. The result is denoted by $m_i =m(c_i)$. The $m_i$ is not necessarily a singleton. We will somewhat abuse the notation by denoting $m_i$ a single or any of the UMLS terms obtained by mapping $c_i$ to UMLS. 
    \item $m(\cdot,\cdot)$ - function that maps pairs of $c_i$ and $c_j$ into the corresponding set of all possible UMLS pairs $m_i$ and $m_j$. Recall that the mapping of $c_i$ to UMLS may not be unique. In this case $|m(c_i,c_j)| = |m(c_i)|\cdot |m(c_j)|$. 
    \item $(m_i, m_j)$ - a pair of UMLS CUIs, which is extracted as a co-occurrence from MEDLINE records. It also represents an edge in network $G$ and is cross-referenced with biocurated databases; 
    \item $D$ - set of pairs $(m_i, m_j)$ extracted from biocurated databases;
    \item $P$ - set of pairs $(m_i, m_j)$ extracted from MEDLINE abstracts;
    \item $E$ - set of cross-referenced pairs $(m_i, m_j)$, such that $E = D \cap P$ ;
    \item $G$ - dynamic network, containing temporal snapshots $G_t$, where $t$ - timestamp (year);
    \item $\hat{G}_t$ - snapshot of network $G$ for a timestamp $t$ only containing nodes from $G_{t-1}$;
\end{itemize}

The main unit of analysis in HG  is a connection between two biomedical concepts, which we also refer to as "pair", "pairwise interaction" or "edge" (in network science context when we will be discussing semantic networks). These connections can be obtained from two main sources: biomedical databases and scientific texts. 
Extracting pairs from biomedical databases is done with respect to the nature and content of the database: some of them already contain pairwise interactions, whereas others focus on more complex structures such as pathways which may contain multiple pairwise interactions or motifs (e.g.,  KEGG~\cite{10.1093/nar/gkv1070}). Extracting pairs from textual data is done via information retrieval methods, such as relation extraction or co-occurrence mining. In this work, we use the abstract-based co-occurrence approach, which is explained later in the paper.

\subsection{Method in Summary}
\begin{figure}[t]
    \centering
    \includegraphics[width=0.85\linewidth]{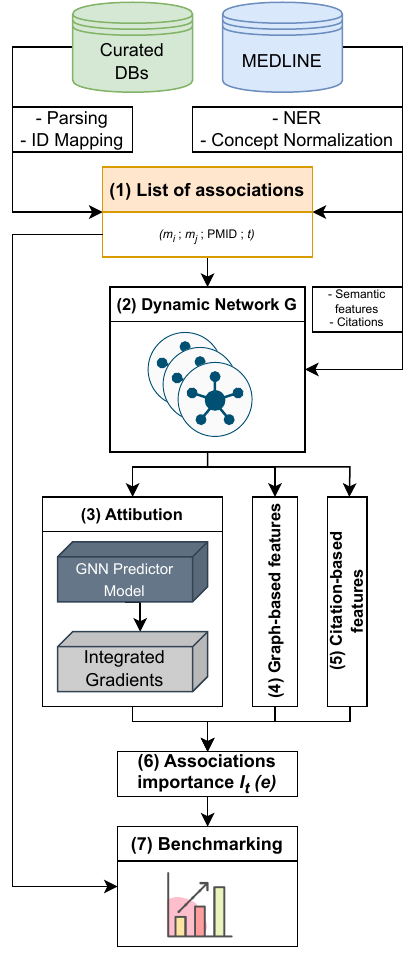}
    \caption{Summary of the HG benchmarking approach. We start with collecting data from Curated DBs and Medline, then process it: records from Curated DBs go through parsing, cleaning and ID mapping, MEDLINE records are fed into SemRep system, which performs NER and concept normalization. After that we obtain a list of UMLS CUI associations with attached PMIDs and timestamps (TS). This data is then used to construct a dynamic network $G$, which is used to calculate the importance measure $I$ for edges in the network. At the end, edges $e \in G$ with their corresponding importance scores $I_t(e)$ are added to the benchmark dataset.}
    \label{fig:bench_pipeline}
\end{figure}
The HG benchmarking pipeline is presented in Figure~\ref{fig:bench_pipeline}. 
The end goal of the pipeline is to provide a way to evaluate any end-to-end hypothesis generation system trained to predict potential pairwise associations between biomedical instances or concepts.

We start with collecting pairwise entity associations from a list of biocurated databases, which we then normalize and represent as pairs of UMLS ~\cite{bodenreider2004unified} terms $(m_i, m_j)$. The set of these associations is then cross-referenced with scientific abstracts extracted from MEDLINE database, such that for each pair $(m_i, m_j)$ we keep all PubMed identifiers (PMID) that correspond to the paper abstracts in which $m_i$ and $m_j$ co-occured. As a result, there is a list of tuples (step 1, Fig. \ref{fig:bench_pipeline}) $(m_i, m_j, \text{PMID}, t)$, where $t$ is a timestamp for a given PMID extracted from its metadata. We then split this list into a sequence  $\{E_t\}$ according to the timestamp $t$. In this work $t$ is taken with a yearly resolution.

Each individual $E_t$ can be treated as an edgelist, which yields an edge-induced network $G_t$ constructed from edges $(m_i, m_j) \in E_t$. It gives us a sequence of networks $G = \{G_t\}$ (step 2, Fig. \ref{fig:bench_pipeline}), which is then used to compute the \textit{importance} of individual associations in  $E_t$ with different methods.

The main goal of importance is to describe each edge from $E_t$ using additional information. The majority of it comes from the future network snapshot $G_{t+1}$, which allows us to track the \textit{impact} that a particular edge had on the network in the future. The predictive impact is calculated with an attribution technique called Integrated Gradients (IG) (step 3, Fig. \ref{fig:bench_pipeline}). Structural impact is calculated with graph-based measures (such as centrality) (step 4, Fig. \ref{fig:bench_pipeline}) and citation impact is calculated with respect to how frequently edges are referenced in the literature after their initial discovery (step 5, Fig. \ref{fig:bench_pipeline}).

All the obtained scores are then merged together to obtain a ranking $I_t(e)$ (step 6, Fig. \ref{fig:bench_pipeline}), where $e \in E_t$ for all edges from a snapshot $G_t$.
Finally, this ranking is used to perform stratified evaluation of how well hypothesis generation systems perform in discovering connections with different importance values (step 7, Fig. \ref{fig:bench_pipeline}).
    
\subsection{Databases Processing and Normalization}
We begin by gathering the links and relationships from publicly available databases, curated by domain experts. 

Ensuring correct correspondence of the same concepts from diverse databases is highly crucial. Therefore, we also conduct mapping of all concepts to UMLS CUI (Concept Unique Identifier).  Concepts, which identifiers cannot be mapped to UMLS CUI, are dropped. In our process, we sometimes encounter situations where a concept $c_{i}$, may have multiple mappings to UMLS CUIs, i.e., $|m_i|=k>1$ for $m_i = m(c_i)$. 
%This can be represented as $|m(c_{i})|  > 1$, where $m(c_{i})$ denotes the set of UMLS mappings for \Ait{$c_{i}: m(c_i) = \{m_{i \rightarrow 1},...,m_{i \rightarrow k}\}$, where $k$ - number of different UMLS terms for $c_i$} and $|\cdot|$ denotes the cardinality or size of the set; \Ait{therefore, $|m(c_i)| = k$.}
To capture these diverse mappings, we use the Cartesian product rule. In this approach, we take the mapping sets for both concepts $c_{i}$ and $c_{j}$, denoted as $m(c_{i})$ and $m(c_{j})$, and generate a new set of pairs encapsulating all possible combinations of these mappings. Essentially, for each original pair $(c_{i}, c_{j})$, we produce a set of pairs $m(c_{i}, c_{j})$ such that the cardinality of this new set equals the product of the cardinalities of the individual mappings. Let us say that $c_i$ has $k$ different UMLS mappings and $c_j$ has $s$, then
$|m(c_{1},c_{2})| = |m(c_{1})| \cdot |m(c_{2})| = k\cdot s$.

In other words, we ensure that every possible mapping of the original pair is accounted for, enabling our system to consider all potential pairwise interactions across all UMLS mappings. To this end, we have collected all pairs of UMLS CUI that are present in different datasets, forming a set $D$.
        
\subsection{Processing MEDLINE Records}

To match pairwise interactions extracted from biocurated databases to literature, we use records from MEDLINE database with their PubMed identifiers.
These records, primarily composed of the titles and abstracts of scientific papers, are each assigned a unique PubMed reference number (PMID). They are also supplemented with rich metadata, which includes information about authors, full-text links (when applicable), and date of publication timestamps indicating when the record became publicly available. 
We process records with an NLM-developed natural language processing tool SemRep~\cite{RINDFLESCH2003462} to perform named entity recognition, concept mapping and normalization. To this end, we obtain a list of UMLS CUI for each MEDLINE record.

\subsection{Connecting Database Records with Literature}
The next step is to form connections between biocurated records and their corresponding mentions in the literature. 
With UMLS CUIs identified in the previous step, we track the instances where these CUIs are mentioned together within the same scientific abstract. Our method considers the simultaneous appearance of a pair of concepts, denoted as $m_i$ and $m_j$, within a single abstract to represent a co-occurrence. This co-occurrence may indicate a potential relationship between the two concepts within the context of the abstract. All the co-occurring pairs $(m_i, m_j)$, extracted from MEDLINE abstracts, form the set $P$.

No specific ``significance''  score is assigned to these co-occurrences at this point beyond their presence in the same abstract. Subsequently, these pairs are cross-referenced with pairs in biocurated databases. More specifically, for each co-occurrence $(m_i, m_j) \in P$ we check its presence in set $D$. Pairs not present in both sets $D$ and $P$ are discarded. This forms the set $E$:
    \begin{equation}
        E = D \cap P.
    \end{equation}
This step validates each co-occurring pair, effectively reducing noise and confirming that each pair holds biological significance. The procedure is described in~\cite{xing2020biorel} as \textit{distant supervised annotation}.
%}
        
\subsection{Constructing Time-sliced Graphs}
After we find the set of co-occurrences in abstracts extracted from MEDLINE and cross-referenced with pairs in biocurated databases (set $E$), we split it based on the timestamps extracted from the abstracts metadata.
The timestamps $t$ are assigned to each PMID and are used to determine when they became publicly available. We use these timestamps to track how often was  a pair of UMLS CUIs $(m_i, m_j)$  appearing in the biomedical literature over time.
As a result, we have a list of biologically relevant cross-referenced UMLS CUI co-occurrences, each connected to all PMIDs containing them.

This list is then split into edge lists $E_t$, such that each edge list contains pairs $(m_i, m_j)$ added in or before year $t$. These edge lists
 are then transformed to dynamic network $G$ with $T$ snapshots:
$$G = \{G_t=(N_t, E_t)\}_{t=1}^T,$$%,\  \forall t \in T,$$
%$$G_{t} = (N_t, E_t),$$
where $N_t$ and $E_t$ represent the set of unique UMLS CUIs (nodes) and their cross-referenced abstract co-occurrences (edges), respectively, and $t$ is the annual timestamp (time resolution can be changed as needed), such that $G_{t}$ is constructed from all MEDLINE records published before $t$ (e.g., $t=2011$). All networks $G_{t}$ are simple and undirected.

For each timestamp $t$, $G_{t}$ represents a \textit{cumulative} network, including all the information from $G_{t-1}$ and new information added in year $t$.
    
\subsection{Tracking the Edge Importance of Time-sliced Graphs}%\added[id=is]{section title is not clear}
We enrich the proposed benchmarking strategy  with the information about associations importance at each time step $t$. In the context of scientific discovery, the importance may be considered from several different perspectives, e.g., as an the influence of an individual finding on future discoveries. In this section we take three different perspectives into account and then combine them together to obtain a final importance score, which we later use to evaluate different hypothesis generation systems with respect to their ability to predict the important associations.

\subsubsection{Integrated Gradients Pipeline}
In this  step we  obtain the information about how edges from graph $G_t$ influence the appearance of new edges in $G_{t+1}$. 
For that we train a machine learning model, which is able to perform link predictions and then we use an attribution method called Integrated Gradients (IG). 
        
In general, IG is used to understand input features importance with respect to the output a given predictor model produces. In case of link prediction problem, a model outputs likelihood of two nodes $m_i$ and $m_j$ being connected for a given network $G_t$. The input features for a link prediction model will include the adjacency matrix of $G_t$, $A_t$, and the predictions themselves can be drawn from a list of edges appearing in the next timestamp $t + 1$. If IG is applied to this particular problem, it will provide attribution values for each element of $A_t$, which can be reformulated as the importance of edges existing at the timestamp $t$ with respect to their contribution to predicting the edges added at the next timestamp $t+1$. This could be interpreted as the \textit{influence} of current dynamic network structural elements on the information that will be added in future.

\paragraph{Link prediction problem} 
In our setting, the link prediction problem is formulated as following:
$$\text{given: } G_{t} = (N_t, E_t)$$
$$\text{predict: } (m_i, m_j) \ \forall m_i, m_j \in \hat{G}_{t+1} (N_t, \hat{E}_{t+1}).$$
\noindent We note that predictions of edges $\hat{E}_{t+1}$ are performed only for nodes $N_t$ from the graph $G_t$ at year $t$.
        
\paragraph{Adding Node and Edge Features}
            
To enrich the dynamic network $G$ with non-redundant information extracted from text, we add node features and edge weights. Node features are required for Graph Neural Network-based predictor training,  which we use in the proposed pipeline.
            
\emph{Node features}:
Node features are added to each  $G_t$ by applying word2vec algorithm~\cite{mikolov2013distributed} to the corresponding snapshot of MEDLINE dataset obtained for a timestamp $t$. 
            In order to perform cleaning and normalization, we replace all tokens in the input texts by their corresponding UMLS CUIs obtained at the NER stage. It significantly reduces the vocabulary size, automatically removing stop-words and enabling vocabulary-guided phrase mining~\cite{aronson2001effective}. It is important to note that each node $m$ has a different vector representation for each time stamp $t$, which we can refer to as $n2v(m, t)$.
            
            \textit{Edge features (weights)}: For simplicity, edge weights are constructed by counting the number of MEDLINE records mentioning a pair of concepts $e \in E_{t}$. In other words, for each pair $e = (m_i, m_j)$ we assign a weight representing the total number of mentions for a pair $e$ in year $t$.
        
\subsubsection{GNN training}
We use a graph neural network-based encoder-decoder architecture. Its encoder consists of two graph convolutional layers~\cite{welling2016semi} and produces an embedding for each graph node. Decoder takes the obtained node embeddings and outputs the sum of element-wise multiplication of encoded node representations as a characteristic of each pair of nodes.
            
\subsubsection{Attribution}
To obtain a connection between newly introduced edges $\hat{E}_{t+1}$ and existing edges $E_t$, we use an attribution method Integrated Gradients (IG)~\cite{sundararajan2017axiomatic}. 
It is based on two key assumptions:
    \begin{itemize}
        \item \textbf{Sensitivity:} any change in input that affects the output gets a non-zero attribution;
        \item \textbf{Implementation Invariance:} attribution is consistent with the model's output, regardless of the model's architecture.
    \end{itemize}
The IG can be applied to a wide variety of ML models as it calculates the attribution scores with respect to input features and not the model weights/activations, which is important, because we focus on relationships between the data points and not the model internal structure.
        
The integrated gradient (IG) score along $i^{th}$ dimension for an input $x$ and baseline $x'$ is defined as:
\begin{equation}
    IG_{i}(x) ::= (x_i - x'_i) \int_{\alpha=0}^{1} \frac{
    \partial F(x' + \alpha  (x-x') )}
    {\partial x_i }
    d\alpha,
\end{equation}
where $\frac{\partial F(x)}{\partial x_i}$ is the gradient of $F(x)$ along $i^{th}$ dimension. In our case, input $x$ is the adjacency matrix of $G_t$ filled with $1s$ as default values (we provide all edges $E_t \in G_t$)  and baseline $x'$ is the matrix of zeroes. As a result, we obtain an adjacency matrix $A(G_t) \in R^{|E_t| \times |E_t|}$ filled with attribution values for each edge $E_t$.

\subsubsection{Graph-based Measures}
    \paragraph{Betweenness Centrality}
        In order to estimate the structural importance of selected edges, we calculate their betweenness centrality~\cite{brandes2001faster}. This importance measure shows the amount of information passing through the edges, therefore indicating their influence over the information flow in the network. It is defined as
        \begin{equation}
            C_B(e) = \sum_{s \neq t \in V} \frac{\sigma_{st}(e)}{\sigma_{st}},
        \end{equation}
        where $\sigma_{st}$ - the number of shortest paths between nodes $s$ and $t$; $\sigma_{st}(e)$ - the number of shortest paths between nodes $s$ and $t$ passing through edge $e$. 
                
        To calculate the betweenness centrality with respect to the future connections, we restrict the set of vertices $V$ to only those, that are involved in future connections we would like to use for explanation. 
    
    \paragraph{Eigenvector Centrality}
Another graph-based structural importance metric we use is the eigenvector centrality. The intuition behind it is that a node  of the network is considered important if it is close to other important nodes. It can be found as a solution of the eigenvalue problem equation:
\begin{equation}
    Ax = \lambda x, 
\end{equation}
where $A$ is the network weighted adjacency matrix. Finding the  eigenvector corresponding to the largest eigenvalue gives us a list of centrality values $C_E(v)$ for each vertex $v \in V$.
        
        However, we are interested in edge-based metric, which we obtain by taking an absolute difference between the adjacent vertex centralities:
        \begin{equation}
            C_E(e) = |C_E(u) - C_E(v) |,
        \end{equation}
        where $e=(u,v)$. The last step is to connect this importance measure to time snapshot, which we do by taking a time-base difference between edge-based eigenvector centralities
        \begin{equation}
            C_{E_{\Delta t}}(e) = C_{E_{t+1}}(e) - C_{E_t}(e),
        \end{equation}
        
        This metric gives us the eigenvector centrality change with respect to future state of the dynamic graph ($t+1$).

    \paragraph{Second Order Jaccard Similarity}
        
        One more indicator of how important a particular newly discovered network connection is related to its adjacent nodes neighborhood similarity. The intuition is that more similar their neighborhood is, more trivial the connection is, therefore, it is less important.

        We consider a second-order Jaccard similarity index for a given pair of nodes $m_i$ and $m_j$:

        \begin{equation}
            J_2(m_i, m_j) = \frac{|N_2(m_i) \cap N_2(m_j)|}{|N_2(m_i) \cup N_2(m_j)|}
        \end{equation}
        Second-order neighborhood of a node $u$ is defined by:

        \begin{equation}
            N_2(u) = \bigcup_{w \in N(u)} N(w),
        \end{equation}
        where $w$ iterates over all neighbors of $u$ and $N(w)$ returns the neighbors of $w$.

        The second order gives a much better "resolution" or granularity for different connections compared to first-order neighborhood. We also note that it is calculated for a graph $G_{t-1}$ for all edges $\hat{E}_{t}$ (before these edges were discovered).

\subsubsection{Literature-based Measures}
    \paragraph{Cumulative citation counts}
    Another measure of a connection importance is related to bibliometrics. At each moment in time for each targeted edge we can obtain a list of papers mentioning this edge.
        
    We also have access to a directed citation network, where nodes represent documents and edges represent citations: edges connect one paper to all the papers that it cites. Therefore, the number of citations of a specific paper would equal to in-degree of a corresponding node in a citation network.
            
To connect paper citations to concepts connections, we compute the sum of citation counts of all papers mentioning a specific connection. This measure shows the overall citation-based impact of a specific edge over time. The citation information comes from the citation graph, which is consistent with the proposed dynamic network in terms of time slicing methodology.

\subsubsection{Combined importance measure for ranking connections}

To connect all the components of the importance measure $I$ for edge $e$, we use the mean percentile rank (PCTRank) of each individual metric:       
%            \Dit{$I_t(e) = \frac{1}{N}\sum_N{\text{PCT rank}(M_t(e))},$}
 \begin{equation}\label{eq:importance}
 I_t(e) = \frac{1}{|M|}\sum_{C_i \in M }{\text{PCTRank}(C_{i_t}(e))},
 \end{equation}
where $C_i$ is the importance component (one of the described earlier). The importance metric is calculated for each individual edge in graph for each moment in time $t$ with respect to its future (or previous) state $t+1$ (or $t-1$). Using the mean percentile rank guarantees that the metric will stay within a unit interval. 
The measure $I$ is used to implement an importance-based stratification strategy for benchmarking, as it is discussed in Results section.
\section{Results}
\begin{figure}
    \centering
    \includegraphics[
        width=1\linewidth,
    ]{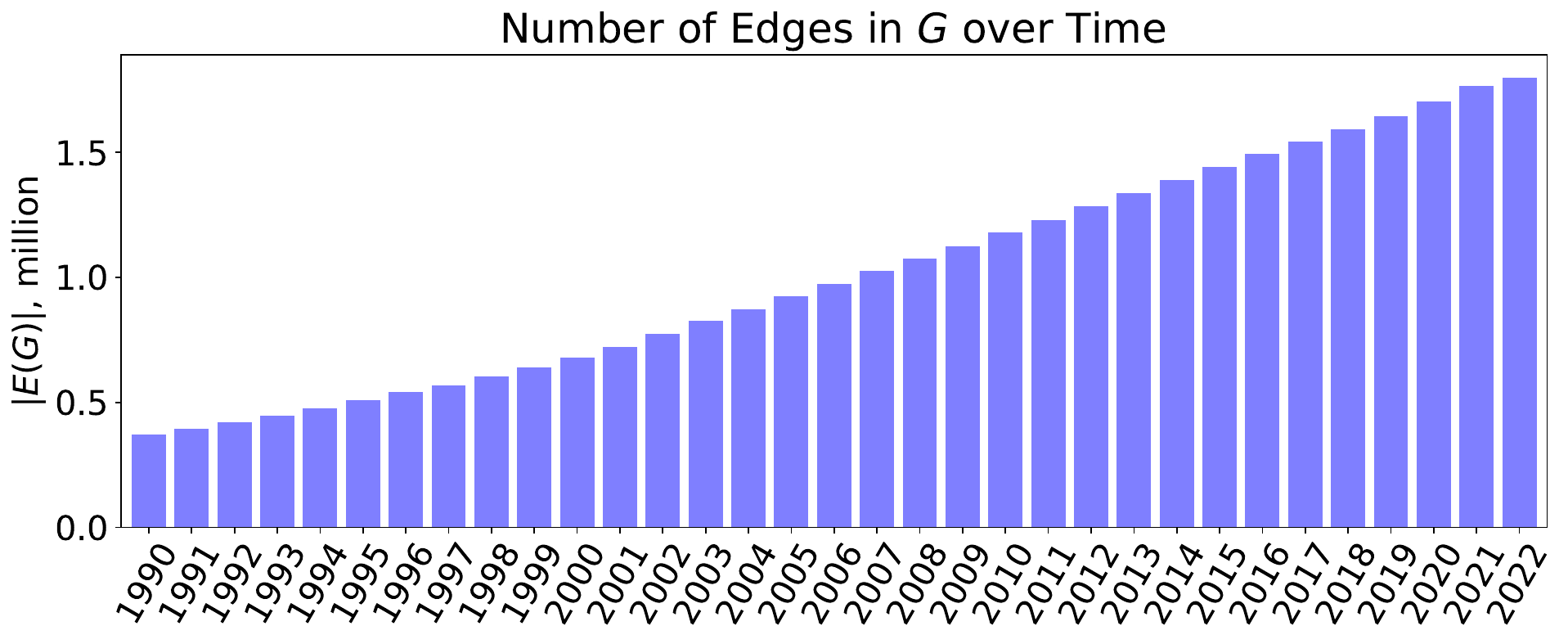}
    \caption{
      \label{fig:res_n_edges_over_time}
      Number of edges in the network $G$ over time. The numbers are reported in millions. Each edge represents a pair of cross-referenced UMLS CUI concepts $(m_i, m_j)$.
    }
\end{figure}
In this section we describe the experimental setup and propose a methodology based on different stratification methods. This methodology is unique for the proposed benchmark, because each record is supplied with additional information giving a user more flexible evaluation protocol.  
    
\subsection{Data Collection and Processing}
\subsubsection{Dynamic Graph Construction}

\begin{table}
    \centering
    \setlength{\tabcolsep}{.35em}
    \begin{tabular}{lrrl}
        \toprule
        {} &  Pairs &  Concepts &                      Concept Types \\
        Database    &                     &                 &                                    \\
        \midrule
        KEGG        &              730214 &           22306 &  Genes, Diseases, Chemicals \\
        CTD         &              459229 &           37065 &         Genes, Diseases, Chemicals \\
        DisGenNET   &              274320 &           18923 &                    Genes, Diseases \\
        DrugCentral &              230805 &           19066 &             Genes, Diseases, Chemicals \\
        RxNav       &              175186 &            2804 &                              Chemicals \\
        STRING      &               63904 &            9118 &                           Proteins \\
        Mentha      &               43673 &           10096 &                           Proteins \\
        GWAS        &               30350 &            8905 &                    Genes, Diseases \\
        \bottomrule
    \end{tabular}
    \caption{
        Experimental databases included in the dataset. Reported numbers are for cross-referenced pairs of UMLS CUI concepts (edges) and the corresponding nodes from the network $G$. 
    }
    \label{tab:bench_sources}    
\end{table}

The numbers of concepts and their associations successfully mapped to UMLS CUI $(m_i, m_j)$ from each dataset are summarized in Table~\ref{tab:bench_sources}. The number of associations with respect to time is shown in Figure~\ref{fig:res_n_edges_over_time}. It can be seen that the number of concept associations steadily and consistently grows for every subsequent year.  

Data collection and aggregation is performed in the following pipeline:
\begin{enumerate}
\item All databases are downloaded in their corresponding formats such as comma-separated or Excel spreadsheets, SQL databases or Docker images.
\item All pairwise interactions in each database are identified. 
\item From all these interactions we create a set of unique concepts, which we then map to UMLS CUIs. Concepts that do not have UMLS representations are dropped.
\item All original pairwise interactions are mapped  with respect to the UMLS codes, as discussed in \textit{Databases Processing and Normalization} section.
\item A set of all pairwise interactions is created by merging the mapped interactions from all databases.
\item\label{pubmed_occur} This set is then used to find pairwise occurrences in MEDLINE. 
\end{enumerate}
Pairwise occurrences found in step \ref{pubmed_occur} are used to construct the main dynamic network $G$. Other pairwise interactions, which are successfully mapped to UMLS CUI, but are not found in the literature, can still be used. They do not have easily identifiable connections to scientific literature and do not contain temporal information, which make them a more difficult target to predict (will be discussed later).
            
\subsubsection{Compound Importance Calculation}

    \begin{table}
    \centering
    \begin{tabular}{lrrrrrr}
        \toprule
        {} &     IG &     EC &     BC &    JC2 &  Ment. &   Cit. \\
        \midrule
        IG    &  1.000 &  0.162 &  0.045 &  0.102 &  0.022 &  0.154 \\
        EC    &  0.162 &  1.000 &  0.032 & -0.028 &  0.091 &  0.119 \\
        BC    &  0.045 &  0.032 &  1.000 & -0.010 & -0.001 &  0.014 \\
        JC2   &  0.102 & -0.028 & -0.010 &  1.000 & -0.050 &  0.018 \\
        Ment. &  0.022 &  0.091 & -0.001 & -0.050 &  1.000 &  0.476 \\
        Cit.  &  0.154 &  0.119 &  0.014 &  0.018 &  0.476 &  1.000 \\
        \bottomrule
    \end{tabular}
    \caption{
          Correlation between components of the proposed importance measure. Used abbreviations: IG - Integrated Gradients; EC - Eigenvector Centrality; BC - Betweenness Centrality; JC2 - 2nd order Jaccard Coefficient (negative); Ment. - Number of mentions; Cit. - Number of citations.
    }
    \label{tab:imp_comp_corr_table}
\end{table}

    Once the dynamic graph $G$ is constructed, we calculate the importance measure. For that we need to decide on three different  timestamps:
    \begin{enumerate}
        \item\label{train_ts} Training timestamp: when the predictor models of interest are trained;
        \item\label{test_ts} Testing timestamp: what moment in time to use to accumulate recently (with respect to step~\ref{train_ts}) discovered concept associations for models testing;
        \item\label{explain_ts} Importance timestamp: what moment in time to use to calculate the importance measure for concept associations from step~\ref{test_ts}.
    \end{enumerate}

To demonstrate our benchmark, we experiment with different predictive models. In our experimental setup, all models are trained on the data published prior to 2016, tested on associations discovered in 2016 and the importance measure $I$ is calculated based on the most recent fully available timestamp (2022) with respect to the PubMed annual baseline release.

The importance measure $I$ has multiple components, which are described in Methods section. To investigate their relationships and how they are connected to each other, we plot a Spearman correlation matrix showed on Figure~\ref{tab:imp_comp_corr_table}. Spearman correlation is used because only component's \textit{rank} matters in the proposed measure as all components are initially scaled differently. 
    
\subsection{Evaluation Protocol}

In our experiments, we demonstrate a scenario for benchmarking hypothesis generation systems. All of the systems are treated as predictors capable of ranking true positive samples (which come from the dynamic network $G$) higher than the synthetically generated negatives. The hypothesis generation problem is formulated as binary classification with significant class imbalance.
        
\subsubsection{Evaluation Metric}

The evaluation metric of choice for our benchmarking is Receiver Operating Characteristic (ROC) curve and its associated Area Under the Curve (AUC), which is calulated as:
            
\begin{equation}
                AUC(f) = \frac{1}{|D^0| \cdot |D^1|} \sum_{t_0 \in D^0} \sum_{t_1 \in D^1} \textbf{1}[f(t_0) < f(t_1)]
            \end{equation}
            where \( \textbf{1} \) is the indicator function that equals 1 if the score of a negative example \( t_0 \) is less than the score of a positive example \( t_1 \); \( D^0 \), \( D^1 \) are the sets of negative and positive examples, respectively. The ROC AUC score quantifies the model's ability to rank a random positive higher than a random negative.
            
            We note than the scores don't have to be within a specific range, the only requirement is that they can be compared with each other. In fact, using this metric allows us to compare purely classification-based models (such as Node2Vec logistic regression pipeline) and ranking models (like TransE or DistMult), even though the scores of these models may have arbitrary values.
        
        \subsubsection{Negative Sampling}

            Our original evaluation protocol can be found in~\cite{sybrandt2020agatha}, which is called \textit{subdomain recommendation}. It is inspired by how the biomedical experts perform large-scale experiments to identity the biological instances of interest from the large pool of candidates~\cite{aksenova2019inhibition}. To summarize:
            \begin{itemize}
                \item We collect all positive samples after a pre-defined cut date. The data before this cut date is used for prediction system training.
                \item For each positive sample (subject-object pair) we generate $N$ negative pairs, such that the subject is the same and the object in every newly generated pair has the same UMLS semantic type as the object in positive pair;
                \item We evaluate a selected performance measure (ROC AUC) with respect to pairs of semantic types (for example, gene-gene or drug-disease) to better understand domain specific differences.
            \end{itemize}
            For this experiment we set $N=10$  as a trade-off between the evaluation quality and runtime. It can be set higher if more thorough evaluation is needed. 
        
    \subsection{Baseline Models Description}

        To demonstrate how the proposed benchmark can be used to evaluate and compare different hypothesis generation system, we use a set of existing models. To make the comparison more fair, all of them are trained on the same snapshots of MEDLINE dataset.
        
        \subsubsection{AGATHA}
        The AGATHA is a general purpose HG system \cite{sybrandt2020agatha,tyagin2022accelerating}  incorporates a multi-step pipeline, which processes the entire MEDLINE database of scientific abstracts, constructs a semantic graph from it and trains a predictor model based on transformer encoder architecture. Besides, the algorithmic pipeline, the key difference between AGATHA and other link prediction systems is that AGATHA is an end-to-end hypothesis generation framework, where the link prediction is only one of its components. 

        \subsubsection{Node2Vec}
        Node2Vec-based predictor is trained as suggested in the original publication~\cite{10.1145/2939672.2939754}. We use a network purely constructed with text-mining-based methods.
        \subsubsection{Knowledge Graph Embeddings-based models}
        Knowledge Graph Embeddings (KGE) models are becoming increasingly popular these days, therefore we include them into our comparison. We use Ampligraph~\cite{ampligraph} library to train and query a list of KGE models: TransE, HolE, ComplEx and DistMult. 
        
    \subsection{Evaluation with Different Stratification}
        \iffalse
            \begin{figure*}
                \centering
                \includegraphics[
                    width=1\linewidth,
                    %height=3cm,
                ]{images/roc_auc_comp_v02.jpg}
                \caption{
                  \label{fig:roc_auc_comp}
                  ROC AUC scores for different models trained on the same PubMed snapshot from 2015 and tested on semantic predicates added in 2016.  
                }
            \end{figure*}
        \fi
        
        \begin{figure}
            \centering
            \includegraphics[
                width=1\linewidth,
            ]{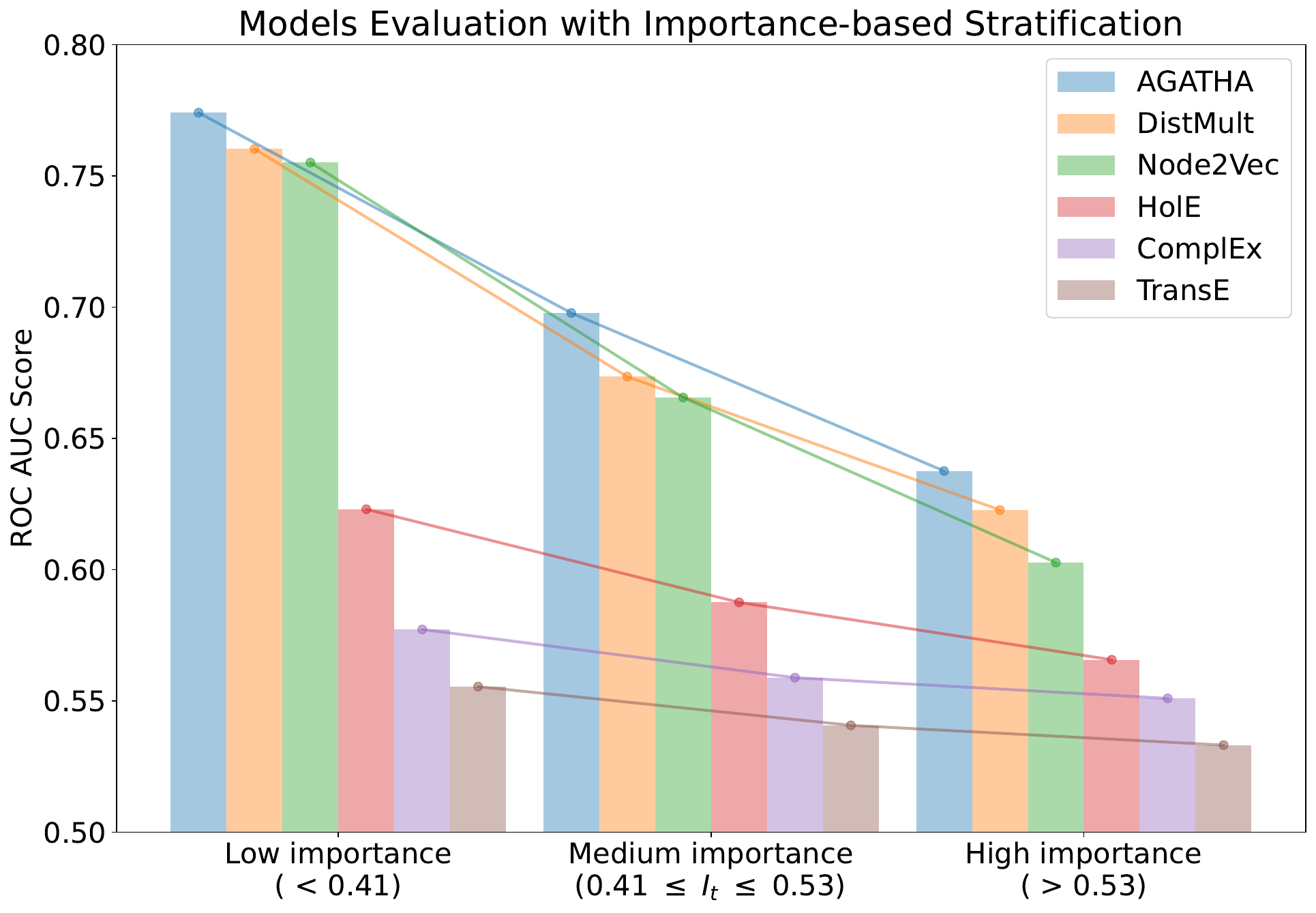}
            \caption{
              \label{fig:res_imp_stratification}
              ROC AUC scores for different models trained on the same PubMed snapshot from 2015 and tested on semantic predicates added in 2016 binned with respect to their importance scores. 
            }
        \end{figure}

        \begin{figure}
            \centering
            \includegraphics[
                width=1\linewidth,
            ]
            %{images/res_temp_stratification.png}
            {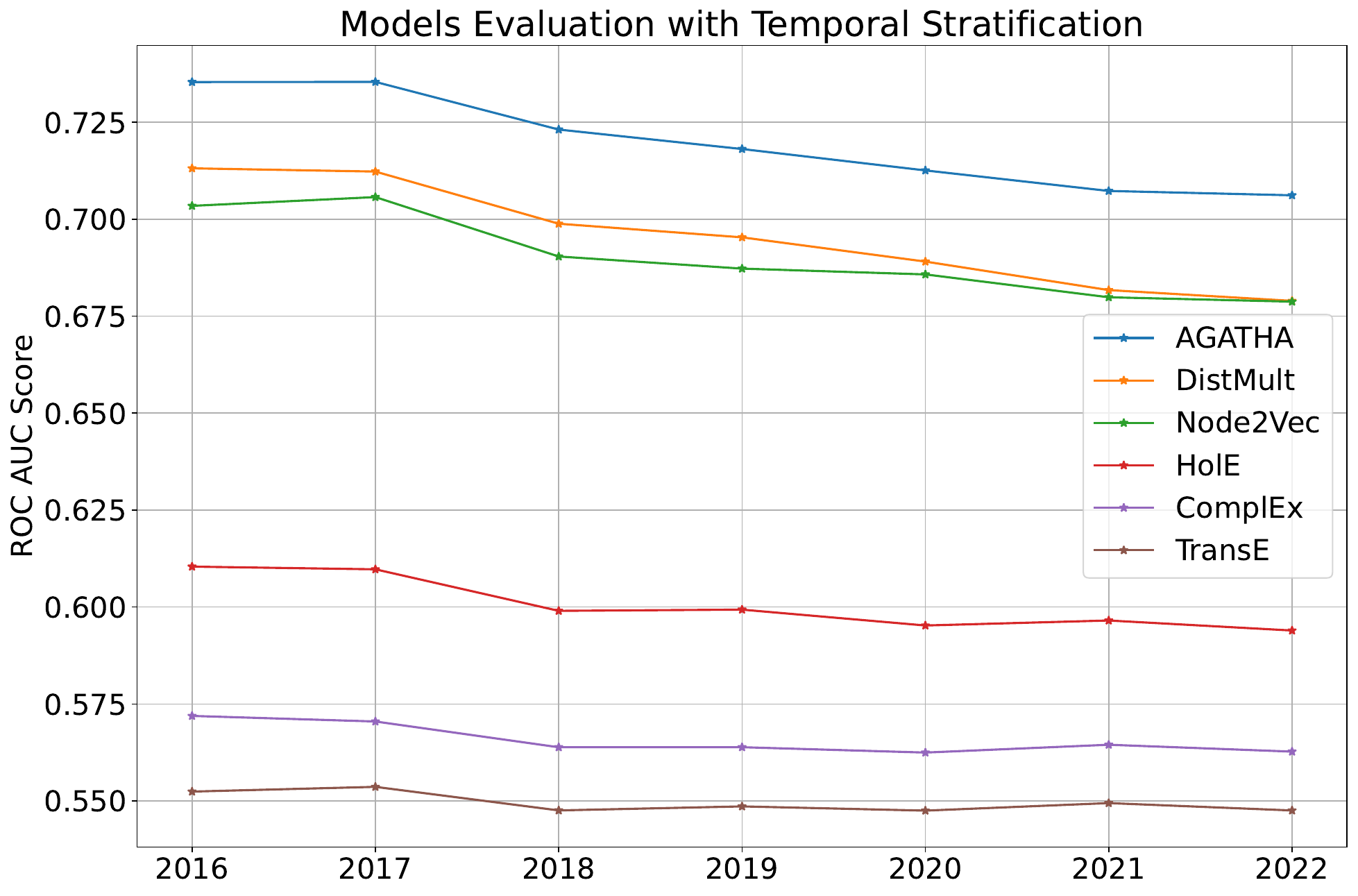}
            \caption{
              \label{fig:res_temp_stratification}
              ROC AUC scores for different models trained on the same PubMed snapshot from 2015 and tested on semantic predicates added over time. 
            }
        \end{figure}

        The proposed benchmarking pipeline enables us to perform different kinds of systems evaluation and comparison with flexibility usually unavailable to other methods. Incorporating both temporal and importance information is helpful to identify trends in models behavior and extend the variety of criteria for domain experts when they decide on a best model suitable for their needs. 

        Below we present three distinct stratification methods and show how predictor models perform under different evaluation protocols. Even though, we use the same performance metric (ROC AUC) across the board, the results differ substantially, suggesting that evaluation strategy plays a significant role in the experimental design. 

\subsubsection{Semantic Stratification}

\begin{table*}[ht]
\small
    \centering
    \begin{tabular}{lrrrrrrr}
    \toprule
    {} &  AGATHA &  DistMult &  Node2Vec &   HolE &  ComplEx &  TransE &  Dataset  \\
    Semantic Pair                                        &         &           &           &        &          &         &         size      \\
    \midrule
    Gene or Genome $\leftrightarrow$ Gene or Genome                    &   0.604 &     0.581 &     0.577 &  0.547 &    0.539 &   0.527 &        606573 \\
    Gene or Genome $\leftrightarrow$ Organic Chemical                  &   0.732 &     0.693 &     0.660 &  0.598 &    0.560 &   0.552 &        359436 \\
    Amino Acid, Peptide, or Protein $\leftrightarrow$ Gene or Genome   &   0.685 &     0.646 &     0.627 &  0.583 &    0.560 &   0.556 &        196493 \\
    Organic Chemical $\leftrightarrow$ Organic Chemical                &   0.918 &     0.897 &     0.893 &  0.679 &    0.617 &   0.589 &        196218 \\
    Gene or Genome $\leftrightarrow$ Pharmacologic Substance           &   0.701 &     0.672 &     0.643 &  0.583 &    0.550 &   0.539 &        109604 \\
    Disease or Syndrome $\leftrightarrow$ Organic Chemical             &   0.876 &     0.871 &     0.855 &  0.673 &    0.607 &   0.566 &        100111 \\
    Amino Acid, Peptide, or Protein $\leftrightarrow$ Organic Chemical &   0.807 &     0.781 &     0.760 &  0.654 &    0.598 &   0.589 &         78111 \\
    Organic Chemical $\leftrightarrow$ Pharmacologic Substance         &   0.890 &     0.870 &     0.871 &  0.656 &    0.590 &   0.565 &         54340 \\
    Disease or Syndrome $\leftrightarrow$ Disease or Syndrome          &   0.833 &     0.826 &     0.822 &  0.636 &    0.574 &   0.555 &         38467 \\
    Disease or Syndrome $\leftrightarrow$ Gene or Genome               &   0.680 &     0.678 &     0.612 &  0.570 &    0.547 &   0.526 &         32549 \\
    \bottomrule
    \end{tabular}
    \caption{
        ROC AUC scores for different models trained on the same MEDLINE snapshot from 2015 and tested on semantic predicates added in the time frame between 2016 and 2022.
    }
    \label{tab:eval_st_strat_table}
    \normalsize
\end{table*}

Semantic stratification strategy is the natural way to benchmark hypothesis generation systems, when the goal is to evaluate performance in specific semantic categories. It is especially relevant to the subdomain recommendation problem, which defines our negative sampling procedure. For that we take the testing set of subject-object pairs and group them according to their semantic types and evaluate each group separately (Table~\ref{tab:eval_st_strat_table}). 
            
\subsubsection{Importance-based Stratification}
        
The next strategy is based on the proposed importance measure $I$. This measure ranks all the positive subject-object pairs from the test set and, therefore, can be used to split them into equally-sized bins, according to their importance score. In our experiment, we split the records into three bins, representing low, medium and high importance values. Negative samples are split accordingly. Then each group is evaluated separately. The results of this evaluation are present in Figure~\ref{fig:res_imp_stratification}.

The results indicate that the importance score $I$ could also reflect the \textit{difficulty} of making a prediction. Specifically, pairs that receive higher importance scores tend to be more challenging for the systems to be identified correctly. In models that generally exhibit high performance (e.g., DistMult), the gap in ROC AUC scores between pairs with low importance scores and those with high importance scores is especially pronounced.

\subsubsection{Temporal Stratification}

The last strategy shows how different models trained once perform \textit{over time}. For that we fix the training timestamp on 2015 and evaluate each models on testing timestamps from 2016 to 2022. For clarity, we do not use importance values for this experiment and only focus on how the models perform over time \textit{on average}. The results are shown in Figure~\ref{fig:res_temp_stratification}.

Figure~\ref{fig:res_temp_stratification} highlights how predictive performance gradually decays over time for every model in the list. This behavior can be expected: the gap between training and testing data increases over time, which makes it more difficult for models to perform well as time goes by. Therefore, it is a good idea to keep the predictor models up-to-date, which we additionally discuss in the next section. 
            
\section{Discussion}
    We divide the discussion into two parts: topics related to evaluation challenges and topics related to different predictor model features.
    
    \subsection{Evaluation-based Topics}
        \subsubsection{Data Collection and Processing Challenges}
        % table with sources 
            The main challenge of this work comes form the diverse nature of biomedical data. This data may be described in many different ways and natural language may not be the most commonly used.
            Our results indicate that a very significant part of biocurated connections "flies under the radar" of text-mining systems and pipelines due to several reasons:
            \begin{enumerate}
                \item Imperfections of text-mining methods;
                \item Multiple standards to describe biomedical concepts;
                \item The diversity of scientific language: many biomedical associations (e.g. gene-gene interactions may be primarily described in terms of co-expression);
                \item Abstracts are not enough for text mining~\cite{sybrandt2018b}.
            \end{enumerate}
            The proposed methodology for the most part takes the lowest common denominator approach: we discard concepts not having UMLS representations and associations not appearing in PubMed abstracts. However, our approach still allows us to extract a significant number of concept associations and to use them for quantitative analysis.
            We should also admit that the aforementioned phenomenon of biomedical data discrepancy leads us to some interesting results, which we discuss below.
            
        \subsubsection{Different Nature of Biomedical DBs and Literature-extracted Data}

            \begin{table*}
    \centering
    \begin{tabular}{lrrrr}
        \toprule
        {} &  Text Mining &  Benchmark &  Non-cross-ref DBs &  Dataset size \\
        Semantic Pair                                                      &              &            &                    &               \\
        \midrule
        Gene or Genome $\leftrightarrow$ Gene or Genome                    &        0.858 &      0.612 &              0.530 &         42625 \\
        Gene or Genome $\leftrightarrow$ Organic Chemical                  &        0.910 &      0.733 &              0.575 &         27060 \\
        Organic Chemical $\leftrightarrow$ Organic Chemical                &        0.905 &      0.922 &              0.679 &         15081 \\
        Amino Acid, Peptide, or Protein $\leftrightarrow$ Gene or Genome   &        0.897 &      0.695 &              0.591 &         14542 \\
        Gene or Genome $\leftrightarrow$ Pharmacologic Substance           &        0.901 &      0.702 &              0.592 &          7843 \\
        Disease or Syndrome $\leftrightarrow$ Organic Chemical             &        0.900 &      0.856 &              0.660 &          7612 \\
        Amino Acid, Peptide, or Protein $\leftrightarrow$ Organic Chemical &        0.906 &      0.820 &              0.564 &          6072 \\
        Organic Chemical $\leftrightarrow$ Pharmacologic Substance         &        0.898 &      0.890 &              0.616 &          4070 \\
        Disease or Syndrome $\leftrightarrow$ Disease or Syndrome          &        0.853 &      0.854 &              0.666 &          2893 \\
        Disease or Syndrome $\leftrightarrow$ Gene or Genome               &        0.847 &      0.690 &              0.575 &          2442 \\
        Non-stratified ROC AUC                                             &        0.886 &      0.715 &              0.579 &        130240 \\
        \bottomrule
    \end{tabular}
    \caption{
        AGATHA-2015 model performance (ROC AUC) evaluated on different data sources with the same cut-off date (where possible). Database records lacking literature cross-references (column 3) were randomly selected because temporal information for them is not available. 
    }
    \label{tab:disc_tm_vs_bench_vs_noncrossref_table}
\end{table*}
            
            The experiment clearly indicates significant differences between different kinds of associations with respect their corresponding data sources in models performance comparison. For this experiment we take one of the evaluated earlier systems (AGATHA 2015) and run the semantically-stratified version of benchmark collected from three different data sources:
            \begin{enumerate}
                \item Proposed benchmark dataset: concept associations extracted from biocurated databases with cross-referenced literature data;
                \item Concept associations extracted from biocurated databases, but which we could not cross-reference with literature data;
                \item Dataset composed of associations extracted with a text mining framework (SemRep).
            \end{enumerate}
            
            Datasets (1) and (3) were constructed from associations found in MEDLINE snapshot from 2020. For dataset (2) it was impossible to identify the time connections were added, therefore the cut date approach was not used. All three datasets were downsampled with respect to the proposed benchmark (1), such that the number of associations is the same across all of them.
            
            The results of this experiment are shown in Table~\ref{tab:disc_tm_vs_bench_vs_noncrossref_table}. 
            It is evident that associations extracted from biocurated databases (1) and (2) propose a more significant challenge for a text-mining-based system. Cross-referencing from literature makes sure that similar associations can be at least discovered by these systems at the training time, therefore, the AGATHA performance on dataset (1) is higher compared to dataset (2).
            These results may indicate that biocurated associations, which cannot be cross-referenced, belong to a different data distribution, and, therefore, purely text mining-based systems fall short due to the limitations of the underlying information extraction algorithms. 
            
    \subsection{Models-related Topics}
        \subsubsection{Text Mining Data Characteristics}

            \begin{figure*}[ht]
            \centering
            % Subfigure A
            \begin{subfigure}[b]{1\linewidth} % Adjust the width as needed
                \includegraphics[width=\linewidth]{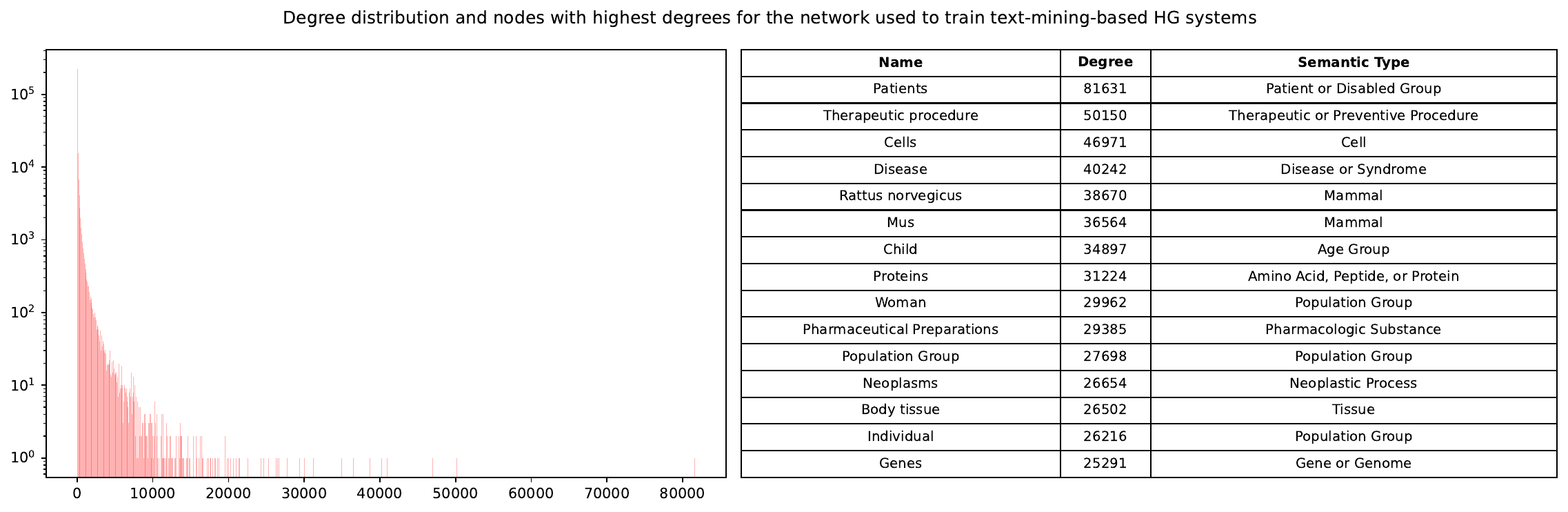}
                %\caption{Caption for subfigure A}
                \label{fig:sub1}
            \end{subfigure}
            \hspace{0.1cm} % Space between the subfigures
            % Subfigure B
            \begin{subfigure}[b]{1\linewidth} % Adjust the width as needed
                \includegraphics[width=\linewidth]{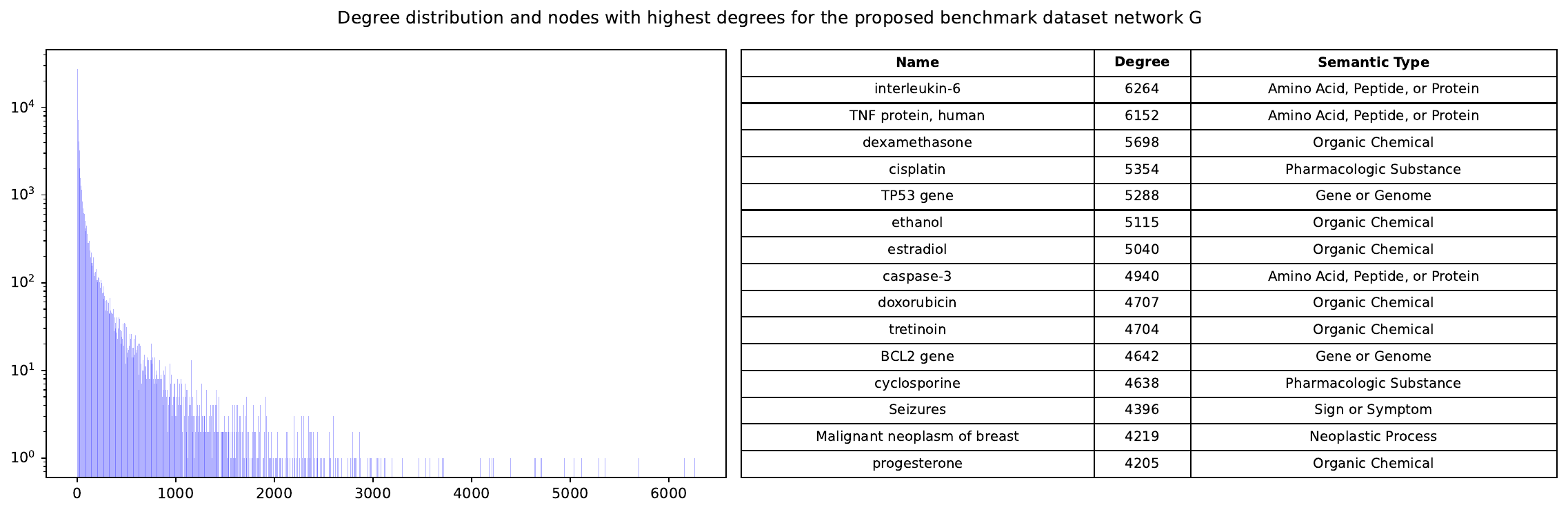}
                %\caption{Caption for subfigure B}
                \label{fig:sub2}
            \end{subfigure}
            \caption{
                Degree distributions and nodes with highest degrees for two networks: the one used for training of text-mining-based predictor models (red, top) and the network $G$ from the proposed benchmark dataset (blue, bottom).
            }
            \label{fig:disc_tm_vs_db_graph_degr_plot_df}
        \end{figure*}
        
            In order to demonstrate the differences between biologically curated and text mining-based knowledge, we can consider their network representations. 
        
            The network-based models we show in this work are trained on text-mining-based networks, which are built on top of semantic predicates extracted from a NLP tool SemRep. This tool takes biomedical text as input and extracts triples \textit{(subject-verb-object)} from the text and performs a number of additional tasks, such as:
            \begin{itemize}
                \item Named Entity Recognition
                \item Concept Normalization
                \item Co-reference Resolution
            \end{itemize}
            and some others. This tool operates on UMLS Metathesaurus, one of the largest and most diverse biomedical thesaurus, including many different vocabularies.
            
            The main problem of text-mining tools like SemRep is that they tend to produce noisy (and often not quite meaningful from the biomedical prospective) data. As a result, the underlying data that is used to build and validate literature-based discovery systems may not represent the results that domain experts expect to see.
            
            However, these systems are automated and, therefore, are widely used as a tool to extract information from literature in uninterrupted manner. Then this information is used for training different kinds of predictors (either rule-based, statistical or deep learning).
            
            To demonstrate this phenomenon, we compare two networks, where nodes are biomedical terms and edges are associations between them. The difference between them lies in their original data source, which is either:
            \begin{enumerate}
                \item PubMed abstracts processed with SemRep tool;
                \item Biocurated databases, which connections are mapped to pairs of UMLS CUI terms and cross-referenced with MEDLINE records.
            \end{enumerate}
            Connections from the network (2) are used in the main proposed benchmarking framework (network $G$). The comparison is shown in Figure~\ref{fig:disc_tm_vs_db_graph_degr_plot_df} as a degree distribution of both networks. We can see that network (1) has a small number of very high-degree nodes. These nodes may affect negatively to the overall predictive power of any model using networks like (1) as a training set, because they introduce a large number of "shortcuts" to the network, which do not have any significant biological value. 
            We also show the top most high-degree nodes for both networks. For the network (1), \textit{all} of them appear to be very general and most of them (e.g. "Patients" or "Pharmaceutical Preparations") can be described as noise. Network (2), in comparison, contain real biomedical entities, which carry domain-specific meaning.

        \subsubsection{Training Data Threshold Influence}

\begin{table}
\small
    \centering
    \setlength{\tabcolsep}{.3em}
    \begin{tabular}{lrr}
        \toprule
        {} &   A-15 &   A-20 \\
        Semantic Pair                                                      &        &        \\
        \midrule
        Amino Acid, Peptide, or Protein $\leftrightarrow$ Gene or Genome   &  0.676 &  0.676 \\
        Amino Acid, Peptide, or Protein $\leftrightarrow$ Organic Chemical &  0.772 &  0.774 \\
        Disease or Syndrome $\leftrightarrow$ Disease or Syndrome          &  0.838 &  0.849 \\
        Disease or Syndrome $\leftrightarrow$ Neoplastic Process           &  0.831 &  0.838 \\
        Disease or Syndrome $\leftrightarrow$ Organic Chemical             &  0.872 &  0.884 \\
        Gene or Genome $\leftrightarrow$ Gene or Genome                    &  0.600 &  0.605 \\
        Gene or Genome $\leftrightarrow$ Organic Chemical                  &  0.718 &  0.725 \\
        Gene or Genome $\leftrightarrow$ Pharmacologic Substance           &  0.691 &  0.692 \\
        Organic Chemical $\leftrightarrow$ Organic Chemical                &  0.907 &  0.913 \\
        Organic Chemical $\leftrightarrow$ Pharmacologic Substance         &  0.908 &  0.916 \\
        Non-stratified ROC AUC                                             &  0.691 &  0.695 \\
        \bottomrule
    \end{tabular}
    \caption{
        ROC AUC comparison between two AGATHA models trained on different MEDLINE snapshots: 2015 and 2020. A-15(20) stands for AGATHA 2015(20).
    }
    \label{tab:disc_a15_vs_a20_table}
\end{table}
\normalsize
            
            As the Temporal Stratification experiment in the results section suggests, the gap between training and testing timestamps plays a noticeable role in models predictive performance. 
            
            To demonstrate this phenomena from a different perspective, we now fix the \textit{testing timestamp} and vary the training timestamp.
            We use two identical AGATHA instances, but trained on different MEDLINE snapshots: 2015 and 2020. The testing timestamp for this experiment is 2021, such that none of the models has access to the test data.

            The results shown in Table~\ref{tab:disc_a15_vs_a20_table} illustrate that having more recent training data does not significantly increase model's predictive power for the proposed benchmark. This result may be surprising, but there is a possible explanation: a model learns the patterns from the training data distribution and that data distribution stays consistent for both training cut dates (2015 and 2020). However, that does not mean that the data distribution in the benchmark behaves the same way. In fact, it changes with respect to both data sources: textual and DB-related.
            
        \subsubsection{Semantic Types Role in Predictive Performance}

            \begin{table}
    \centering
    \begin{tabular}{lrrrr}
        \toprule
        {} &  ComplEx &  DistMult &   HolE &  TransE \\
        \midrule
        No ST &    0.564 &     0.680 &  0.595 &   0.549 \\
        ST    &    0.693 &     0.697 &  0.718 &   0.678 \\
        \bottomrule
    \end{tabular}
    \caption{
          ROC AUC scores comparison (non-stratified) between KGE-based models with and without semantic types information (ST) added to the training set.
    }
    \label{tab:kge_st_comp_table}
\end{table}

            Another aspect affecting models predictive performance is having access to domain information. Since we formulate the problem as subdomain recommendation, knowing concept-domain relationships may be particularly valuable. We test this idea by injecting semantic types information into the edge type for tested earlier Knowledge Graph Embedding models. As opposed to classic link prediction methods (like node2vec), Knowledge Graph modeling was designed around typed edges and allows this extension naturally. 

            Results in Table~\ref{tab:kge_st_comp_table} show that semantic type information provides a very significant improvement for models predictive performance.
            
        \subsubsection{Large Language Models for Scientific Discovery}
            % figure with BioGPT confusion matrix 
            \begin{figure}
                \centering
                \includegraphics[
                    width=1\linewidth,
                ]{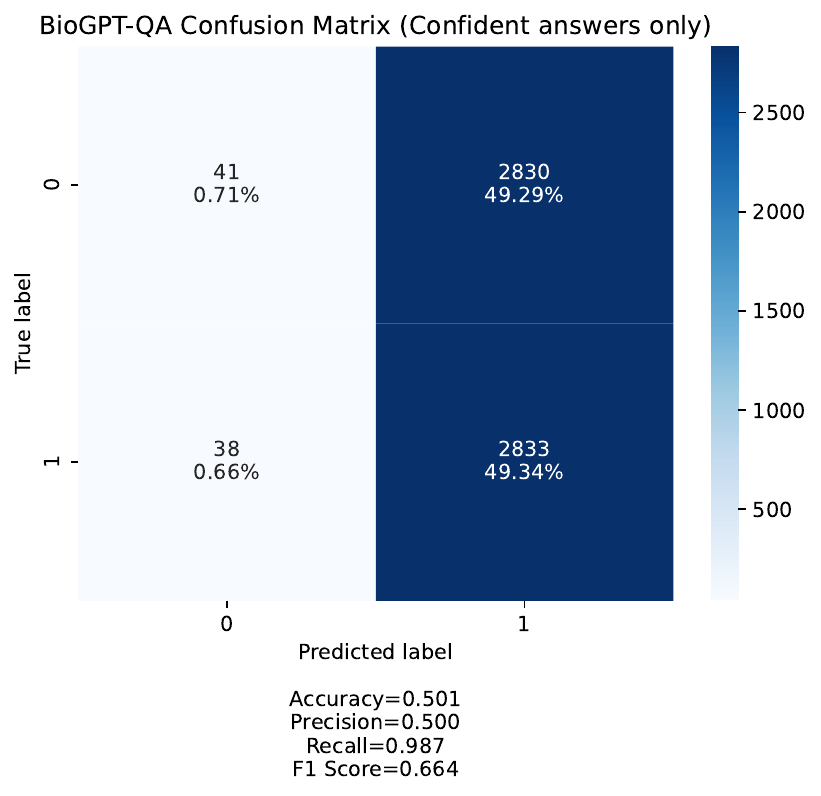}
                \caption{
                  \label{fig:biogpt_conf_matr}
                  Confusion matrix obtained by the BioGPT-QA model. Only confident answers (Yes/No) were taken into account.
                }
            \end{figure}
        
            Recent advances in language model development raised a logical question about usefulness of these models in scientific discovery, especially in biomedical area~\cite{liu2021ai}. Problems like drug discovery, drug repurposing, clinical trial optimization and many others may benefit significantly from systems trained on a large amount of scientific biomedical data.
        
            Therefore, we decide to test how one of these systems would perform in our benchmark.
            We take one of the recently released generative pre-trained transformer models BioGPT~\cite{luo2022biogpt} and run a set of test queries.
            
            BioGPT model was chosen due to the following reasons:
            \begin{itemize}
                \item It is recently released (2022);
                \item It includes fine-tuned models, which show good performance on downstream tasks;
                \item It is open source and easily accessible.
            \end{itemize}
            
            We use a BioGPT-QA model to perform the benchmarking, because it was fine-tuned on PubMedQA~\cite{jin2019pubmedqa} dataset and outputs the answer as yes/maybe/no, which is easy to parse and represent as a (binary) classifier output. 
            
            The question prompt was formulated as the following: "Is there a relationship between term 1 and term 2?". PubMedQA format also requires a context from a PubMed abstract, which does not exist in our case, because it is a discovery problem. However, we supply an abstract-like context, which is constructed by concatenating term definitions extracted from UMLS Metathesaurus for both source and target terms.

            A sample prompt looks like this: \textit{"Is there a relationship between F1-ATPase and pyridoxal phosphate? context: F1-ATPase - The catalytic sector of proton-translocating ATPase complexes. It contains five subunits named alpha, beta, gamma, delta and eta. pyridoxal phosphate - This is the active form of VITAMIN B6 serving as a coenzyme for synthesis of amino acids, neurotransmitters (serotonin, norepinephrine), sphingolipids, aminolevulinic acid..."}
            
            When we ran the experiment, we noticed two things:
            \begin{itemize}
                \item BioGPT is often not confident in its responses, which means that it outputs "maybe" or two answers (both "yes" and "no") for about 40\% of the provided queries;
                \item The overwhelming majority of provided queries are answered positively when the answer is confident.
            \end{itemize}
        
            Figure~\ref{fig:biogpt_conf_matr} shows a confusion matrix for queries with confident answer. We generate the queries set with 1:1 positive to negative ratio. Most of the answers BioGPT-QA provides are positive, which means that the system produces too many false positives and is not usable in the discovery setting. 
\section{Conclusions}

We have developed and implemented a comprehensive benchmarking system \sysname for evaluating biomedical hypothesis generation systems. This benchmarking system is advancing the field by providing a structured and systematic approach to assess the efficacy of various hypothesis generation methodologies.

In our pipeline we utilized several  curated datasets, which provide a basis in testing the hypothesis generation systems under realistic conditions. The informative discoveries have been integrated into the dynamic graph on top of which we introduced the quantification of discovery importance. This approach allowed us to add a new dimension to the benchmarking process, enabling us to not only assess the accuracy of the hypotheses generated but also their relevance and potential impact in the field of biomedical research. This quantification of discovery importance is a critical step forward, as it aligns the benchmarking process more closely with the practical and applied goals of biomedical research.

We have demonstrated the use case of several graph based link prediction systems' verification and concluded that such testing is way more productive than traditional link prediction benchmarks. However, the utility of our benchmarking system extends beyond these examples. We advocate for its widespread adoption to validate the quality of hypothesis generation, aiming to broaden the range of scientific discoveries accessible to the wider research community. Our system is designed to be inclusive, welcoming the addition of more diverse cases.

Future work includes integration of the benchmarking process in the hypothesis system visualization \cite{tyagin2021interpretable}, spreading to other than biomedical areas \cite{marasco2023literature}, integration of novel importance measures, and healthcare benchmarking cases.

\section{Acknowledgements}

This research was supported by NIH award \#R01DA054992. The computational experiments were supported in part through the use of DARWIN computing system: DARWIN - A Resource for Computational and Data-intensive Research at the University of Delaware and in the Delaware Region, which is supported by NSF Grant \#1919839.

%\begin{acks}
%We would like to thank the CCIT staff who manage the Palmetto Supercomputer at Clemson University, where we ran all of our experiments. This work was additionally supported by NSF \#1633608.
%\end{acks}

\bibliographystyle{ACM-Reference-Format}
\bibliography{reference}


\begin{thebibliography}{48}

%%% the Web address.

\ifx \showCODEN    \undefined \def \showCODEN     #1{\unskip}     \fi
\ifx \showDOI      \undefined \def \showDOI       #1{#1}\fi
\ifx \showISBNx    \undefined \def \showISBNx     #1{\unskip}     \fi
\ifx \showISBNxiii \undefined \def \showISBNxiii  #1{\unskip}     \fi
\ifx \showISSN     \undefined \def \showISSN      #1{\unskip}     \fi
\ifx \showLCCN     \undefined \def \showLCCN      #1{\unskip}     \fi
\ifx \shownote     \undefined \def \shownote      #1{#1}          \fi
\ifx \showarticletitle \undefined \def \showarticletitle #1{#1}   \fi
\ifx \showURL      \undefined \def \showURL       {\relax}        \fi
% invisible to TeX
\providecommand\bibfield[2]{#2}
\providecommand\bibinfo[2]{#2}
\providecommand\natexlab[1]{#1}
\providecommand\showeprint[2][]{arXiv:#2}

\bibitem[Aksenova et~al\mbox{.}(2019)]%
        {aksenova2019inhibition}
\bibfield{author}{\bibinfo{person}{Marina Aksenova}, \bibinfo{person}{Justin Sybrandt}, \bibinfo{person}{Biyun Cui}, \bibinfo{person}{Vitali Sikirzhytski}, \bibinfo{person}{Hao Ji}, \bibinfo{person}{Diana Odhiambo}, \bibinfo{person}{Matthew~D Lucius}, \bibinfo{person}{Jill~R Turner}, \bibinfo{person}{Eugenia Broude}, \bibinfo{person}{Edsel Pe{\~n}a}, {et~al\mbox{.}}} \bibinfo{year}{2019}\natexlab{}.
\newblock \showarticletitle{Inhibition of the Dead Box RNA Helicase 3 prevents HIV-1 Tat and cocaine-induced neurotoxicity by targeting microglia activation}.
\newblock \bibinfo{journal}{\emph{Journal of Neuroimmune Pharmacology}} (\bibinfo{year}{2019}), \bibinfo{pages}{1--15}.
\newblock


\bibitem[Aronson(2001)]%
        {aronson2001effective}
\bibfield{author}{\bibinfo{person}{Alan~R Aronson}.} \bibinfo{year}{2001}\natexlab{}.
\newblock \showarticletitle{Effective mapping of biomedical text to the UMLS Metathesaurus: the MetaMap program.}. In \bibinfo{booktitle}{\emph{Proceedings of the AMIA Symposium}}. American Medical Informatics Association, \bibinfo{pages}{17}.
\newblock


\bibitem[Bodenreider(2004)]%
        {bodenreider2004unified}
\bibfield{author}{\bibinfo{person}{Olivier Bodenreider}.} \bibinfo{year}{2004}\natexlab{}.
\newblock \showarticletitle{The unified medical language system (UMLS): integrating biomedical terminology}.
\newblock \bibinfo{journal}{\emph{Nucleic acids research}} \bibinfo{volume}{32}, \bibinfo{number}{suppl\_1} (\bibinfo{year}{2004}), \bibinfo{pages}{D267--D270}.
\newblock


\bibitem[Bonner et~al\mbox{.}(2022a)]%
        {bonner2022review}
\bibfield{author}{\bibinfo{person}{Stephen Bonner}, \bibinfo{person}{Ian~P Barrett}, \bibinfo{person}{Cheng Ye}, \bibinfo{person}{Rowan Swiers}, \bibinfo{person}{Ola Engkvist}, \bibinfo{person}{Andreas Bender}, \bibinfo{person}{Charles~Tapley Hoyt}, {and} \bibinfo{person}{William~L Hamilton}.} \bibinfo{year}{2022}\natexlab{a}.
\newblock \showarticletitle{A review of biomedical datasets relating to drug discovery: a knowledge graph perspective}.
\newblock \bibinfo{journal}{\emph{Briefings in Bioinformatics}} \bibinfo{volume}{23}, \bibinfo{number}{6} (\bibinfo{year}{2022}), \bibinfo{pages}{bbac404}.
\newblock


\bibitem[Bonner et~al\mbox{.}(2022b)]%
        {bonner2022understanding}
\bibfield{author}{\bibinfo{person}{Stephen Bonner}, \bibinfo{person}{Ian~P Barrett}, \bibinfo{person}{Cheng Ye}, \bibinfo{person}{Rowan Swiers}, \bibinfo{person}{Ola Engkvist}, \bibinfo{person}{Charles~Tapley Hoyt}, {and} \bibinfo{person}{William~L Hamilton}.} \bibinfo{year}{2022}\natexlab{b}.
\newblock \showarticletitle{Understanding the performance of knowledge graph embeddings in drug discovery}.
\newblock \bibinfo{journal}{\emph{Artificial Intelligence in the Life Sciences}}  \bibinfo{volume}{2} (\bibinfo{year}{2022}), \bibinfo{pages}{100036}.
\newblock


\bibitem[Bordes et~al\mbox{.}(2013)]%
        {bordes2013translating}
\bibfield{author}{\bibinfo{person}{Antoine Bordes}, \bibinfo{person}{Nicolas Usunier}, \bibinfo{person}{Alberto Garcia-Duran}, \bibinfo{person}{Jason Weston}, {and} \bibinfo{person}{Oksana Yakhnenko}.} \bibinfo{year}{2013}\natexlab{}.
\newblock \showarticletitle{Translating embeddings for modeling multi-relational data}. In \bibinfo{booktitle}{\emph{Advances in neural information processing systems}}. \bibinfo{pages}{2787--2795}.
\newblock


\bibitem[Brandes(2001)]%
        {brandes2001faster}
\bibfield{author}{\bibinfo{person}{Ulrik Brandes}.} \bibinfo{year}{2001}\natexlab{}.
\newblock \showarticletitle{A faster algorithm for betweenness centrality}.
\newblock \bibinfo{journal}{\emph{Journal of mathematical sociology}} \bibinfo{volume}{25}, \bibinfo{number}{2} (\bibinfo{year}{2001}), \bibinfo{pages}{163--177}.
\newblock


\bibitem[Breit et~al\mbox{.}(2020)]%
        {10.1093/bioinformatics/btaa274}
\bibfield{author}{\bibinfo{person}{Anna Breit}, \bibinfo{person}{Simon Ott}, \bibinfo{person}{Asan Agibetov}, {and} \bibinfo{person}{Matthias Samwald}.} \bibinfo{year}{2020}\natexlab{}.
\newblock \showarticletitle{{OpenBioLink: a benchmarking framework for large-scale biomedical link prediction}}.
\newblock \bibinfo{journal}{\emph{Bioinformatics}} \bibinfo{volume}{36}, \bibinfo{number}{13} (\bibinfo{date}{04} \bibinfo{year}{2020}), \bibinfo{pages}{4097--4098}.
\newblock
\showISSN{1367-4803}
\urldef\tempurl%
\url{https://doi.org/10.1093/bioinformatics/btaa274}
\showDOI{\tempurl}
\showeprint{https://academic.oup.com/bioinformatics/article-pdf/36/13/4097/33458980/btaa274\_supplementary\_data.pdf}


\bibitem[Calderone et~al\mbox{.}(2013)]%
        {calderone2013mentha}
\bibfield{author}{\bibinfo{person}{Alberto Calderone}, \bibinfo{person}{Luisa Castagnoli}, {and} \bibinfo{person}{Gianni Cesareni}.} \bibinfo{year}{2013}\natexlab{}.
\newblock \showarticletitle{Mentha: a resource for browsing integrated protein-interaction networks}.
\newblock \bibinfo{journal}{\emph{Nature methods}} \bibinfo{volume}{10}, \bibinfo{number}{8} (\bibinfo{year}{2013}), \bibinfo{pages}{690--691}.
\newblock


\bibitem[Cameron et~al\mbox{.}(2015)]%
        {CAMERON2015141}
\bibfield{author}{\bibinfo{person}{Delroy Cameron}, \bibinfo{person}{Ramakanth Kavuluru}, \bibinfo{person}{Thomas~C. Rindflesch}, \bibinfo{person}{Amit~P. Sheth}, \bibinfo{person}{Krishnaprasad Thirunarayan}, {and} \bibinfo{person}{Olivier Bodenreider}.} \bibinfo{year}{2015}\natexlab{}.
\newblock \showarticletitle{Context-driven automatic subgraph creation for literature-based discovery}.
\newblock \bibinfo{journal}{\emph{Journal of Biomedical Informatics}}  \bibinfo{volume}{54} (\bibinfo{year}{2015}), \bibinfo{pages}{141--157}.
\newblock
\showISSN{1532-0464}
\urldef\tempurl%
\url{https://doi.org/10.1016/j.jbi.2015.01.014}
\showDOI{\tempurl}


\bibitem[Chen et~al\mbox{.}(2016)]%
        {chen2016ibm}
\bibfield{author}{\bibinfo{person}{Ying Chen}, \bibinfo{person}{JD~Elenee Argentinis}, {and} \bibinfo{person}{Griff Weber}.} \bibinfo{year}{2016}\natexlab{}.
\newblock \showarticletitle{IBM Watson: how cognitive computing can be applied to big data challenges in life sciences research}.
\newblock \bibinfo{journal}{\emph{Clinical therapeutics}} \bibinfo{volume}{38}, \bibinfo{number}{4} (\bibinfo{year}{2016}), \bibinfo{pages}{688--701}.
\newblock


\bibitem[Costabello et~al\mbox{.}(2019)]%
        {ampligraph}
\bibfield{author}{\bibinfo{person}{Luca Costabello}, \bibinfo{person}{Alberto Bernardi}, \bibinfo{person}{Adrianna Janik}, \bibinfo{person}{Sumit Pai}, \bibinfo{person}{Chan~Le Van}, \bibinfo{person}{Rory McGrath}, \bibinfo{person}{Nicholas McCarthy}, {and} \bibinfo{person}{Pedro Tabacof}.} \bibinfo{year}{2019}\natexlab{}.
\newblock \bibinfo{title}{{AmpliGraph: a Library for Representation Learning on Knowledge Graphs}}.
\newblock
\newblock
\urldef\tempurl%
\url{https://doi.org/10.5281/zenodo.2595043}
\showDOI{\tempurl}


\bibitem[Davis et~al\mbox{.}(2022)]%
        {davis_wiegers_johnson_sciaky_wiegers_mattingly_2022}
\bibfield{author}{\bibinfo{person}{Allan~Peter Davis}, \bibinfo{person}{Thomas~C. Wiegers}, \bibinfo{person}{Robin~J. Johnson}, \bibinfo{person}{Daniela Sciaky}, \bibinfo{person}{Jolene Wiegers}, {and} \bibinfo{person}{Carolyn~J. Mattingly}.} \bibinfo{year}{2022}\natexlab{}.
\newblock \showarticletitle{Comparative Toxicogenomics Database (CTD): update 2023}.
\newblock \bibinfo{journal}{\emph{NUCLEIC ACIDS RESEARCH}} (\bibinfo{date}{Sep} \bibinfo{year}{2022}).
\newblock
\showISSN{["1362-4962"]}
\urldef\tempurl%
\url{https://doi.org/10.1093/nar/gkac833}
\showDOI{\tempurl}


\bibitem[Fey and Lenssen(2019)]%
        {Fey/Lenssen/2019}
\bibfield{author}{\bibinfo{person}{Matthias Fey} {and} \bibinfo{person}{Jan~E. Lenssen}.} \bibinfo{year}{2019}\natexlab{}.
\newblock \showarticletitle{Fast Graph Representation Learning with {PyTorch Geometric}}. In \bibinfo{booktitle}{\emph{ICLR Workshop on Representation Learning on Graphs and Manifolds}}.
\newblock


\bibitem[Fricke(2018)]%
        {fricke2018semantic}
\bibfield{author}{\bibinfo{person}{Suzanne Fricke}.} \bibinfo{year}{2018}\natexlab{}.
\newblock \showarticletitle{Semantic scholar}.
\newblock \bibinfo{journal}{\emph{Journal of the Medical Library Association: JMLA}} \bibinfo{volume}{106}, \bibinfo{number}{1} (\bibinfo{year}{2018}), \bibinfo{pages}{145}.
\newblock


\bibitem[Gordon and Dumais(1998)]%
        {gordon1998using}
\bibfield{author}{\bibinfo{person}{Michael~D Gordon} {and} \bibinfo{person}{Susan Dumais}.} \bibinfo{year}{1998}\natexlab{}.
\newblock \showarticletitle{Using latent semantic indexing for literature based discovery}.
\newblock \bibinfo{journal}{\emph{Journal of the American Society for Information Science}} \bibinfo{volume}{49}, \bibinfo{number}{8} (\bibinfo{year}{1998}), \bibinfo{pages}{674--685}.
\newblock


\bibitem[Grover and Leskovec(2016)]%
        {10.1145/2939672.2939754}
\bibfield{author}{\bibinfo{person}{Aditya Grover} {and} \bibinfo{person}{Jure Leskovec}.} \bibinfo{year}{2016}\natexlab{}.
\newblock \showarticletitle{Node2vec: Scalable Feature Learning for Networks}. In \bibinfo{booktitle}{\emph{Proceedings of the 22nd ACM SIGKDD International Conference on Knowledge Discovery and Data Mining}} (San Francisco, California, USA) \emph{(\bibinfo{series}{KDD '16})}. \bibinfo{publisher}{Association for Computing Machinery}, \bibinfo{address}{New York, NY, USA}, \bibinfo{pages}{855–864}.
\newblock
\showISBNx{9781450342322}
\urldef\tempurl%
\url{https://doi.org/10.1145/2939672.2939754}
\showDOI{\tempurl}


\bibitem[Hristovski et~al\mbox{.}(2005)]%
        {hristovski2005using}
\bibfield{author}{\bibinfo{person}{Dimitar Hristovski}, \bibinfo{person}{Borut Peterlin}, \bibinfo{person}{Joyce~A Mitchell}, {and} \bibinfo{person}{Susanne~M Humphrey}.} \bibinfo{year}{2005}\natexlab{}.
\newblock \showarticletitle{Using literature-based discovery to identify disease candidate genes}.
\newblock \bibinfo{journal}{\emph{International journal of medical informatics}} \bibinfo{volume}{74}, \bibinfo{number}{2} (\bibinfo{year}{2005}), \bibinfo{pages}{289--298}.
\newblock


\bibitem[Jin et~al\mbox{.}(2019)]%
        {jin2019pubmedqa}
\bibfield{author}{\bibinfo{person}{Qiao Jin}, \bibinfo{person}{Bhuwan Dhingra}, \bibinfo{person}{Zhengping Liu}, \bibinfo{person}{William~W Cohen}, {and} \bibinfo{person}{Xinghua Lu}.} \bibinfo{year}{2019}\natexlab{}.
\newblock \showarticletitle{Pubmedqa: A dataset for biomedical research question answering}.
\newblock \bibinfo{journal}{\emph{arXiv preprint arXiv:1909.06146}} (\bibinfo{year}{2019}).
\newblock


\bibitem[Kanehisa et~al\mbox{.}(2015)]%
        {10.1093/nar/gkv1070}
\bibfield{author}{\bibinfo{person}{Minoru Kanehisa}, \bibinfo{person}{Yoko Sato}, \bibinfo{person}{Masayuki Kawashima}, \bibinfo{person}{Miho Furumichi}, {and} \bibinfo{person}{Mao Tanabe}.} \bibinfo{year}{2015}\natexlab{}.
\newblock \showarticletitle{{KEGG as a reference resource for gene and protein annotation}}.
\newblock \bibinfo{journal}{\emph{Nucleic Acids Research}} \bibinfo{volume}{44}, \bibinfo{number}{D1} (\bibinfo{date}{10} \bibinfo{year}{2015}), \bibinfo{pages}{D457--D462}.
\newblock
\showISSN{0305-1048}
\urldef\tempurl%
\url{https://doi.org/10.1093/nar/gkv1070}
\showDOI{\tempurl}
\showeprint{https://academic.oup.com/nar/article-pdf/44/D1/D457/9482226/gkv1070.pdf}


\bibitem[Kilicoglu et~al\mbox{.}(2012)]%
        {KilicogluSFRR12}
\bibfield{author}{\bibinfo{person}{Halil Kilicoglu}, \bibinfo{person}{Dongwook Shin}, \bibinfo{person}{Marcelo Fiszman}, \bibinfo{person}{Graciela Rosemblat}, {and} \bibinfo{person}{Thomas~C. Rindflesch}.} \bibinfo{year}{2012}\natexlab{}.
\newblock \showarticletitle{SemMedDB: a PubMed-scale repository of biomedical semantic predications.}
\newblock \bibinfo{journal}{\emph{Bioinform.}} \bibinfo{volume}{28}, \bibinfo{number}{23} (\bibinfo{year}{2012}), \bibinfo{pages}{3158--3160}.
\newblock
\urldef\tempurl%
\url{http://dblp.uni-trier.de/db/journals/bioinformatics/bioinformatics28.html#KilicogluSFRR12}
\showURL{%
\tempurl}


\bibitem[Kokhlikyan et~al\mbox{.}(2020)]%
        {kokhlikyan2020captum}
\bibfield{author}{\bibinfo{person}{Narine Kokhlikyan}, \bibinfo{person}{Vivek Miglani}, \bibinfo{person}{Miguel Martin}, \bibinfo{person}{Edward Wang}, \bibinfo{person}{Bilal Alsallakh}, \bibinfo{person}{Jonathan Reynolds}, \bibinfo{person}{Alexander Melnikov}, \bibinfo{person}{Natalia Kliushkina}, \bibinfo{person}{Carlos Araya}, \bibinfo{person}{Siqi Yan}, {and} \bibinfo{person}{Orion Reblitz-Richardson}.} \bibinfo{year}{2020}\natexlab{}.
\newblock \bibinfo{title}{Captum: A unified and generic model interpretability library for PyTorch}.
\newblock
\newblock
\showeprint[arxiv]{2009.07896}~[cs.LG]


\bibitem[Liu et~al\mbox{.}(2021)]%
        {liu2021ai}
\bibfield{author}{\bibinfo{person}{Zhichao Liu}, \bibinfo{person}{Ruth~A Roberts}, \bibinfo{person}{Madhu Lal-Nag}, \bibinfo{person}{Xi Chen}, \bibinfo{person}{Ruili Huang}, {and} \bibinfo{person}{Weida Tong}.} \bibinfo{year}{2021}\natexlab{}.
\newblock \showarticletitle{AI-based language models powering drug discovery and development}.
\newblock \bibinfo{journal}{\emph{Drug Discovery Today}} \bibinfo{volume}{26}, \bibinfo{number}{11} (\bibinfo{year}{2021}), \bibinfo{pages}{2593--2607}.
\newblock


\bibitem[Luo et~al\mbox{.}(2022)]%
        {luo2022biogpt}
\bibfield{author}{\bibinfo{person}{Renqian Luo}, \bibinfo{person}{Liai Sun}, \bibinfo{person}{Yingce Xia}, \bibinfo{person}{Tao Qin}, \bibinfo{person}{Sheng Zhang}, \bibinfo{person}{Hoifung Poon}, {and} \bibinfo{person}{Tie-Yan Liu}.} \bibinfo{year}{2022}\natexlab{}.
\newblock \showarticletitle{BioGPT: generative pre-trained transformer for biomedical text generation and mining}.
\newblock \bibinfo{journal}{\emph{Briefings in Bioinformatics}} \bibinfo{volume}{23}, \bibinfo{number}{6} (\bibinfo{year}{2022}), \bibinfo{pages}{bbac409}.
\newblock


\bibitem[Marasco et~al\mbox{.}(2023)]%
        {marasco2023literature}
\bibfield{author}{\bibinfo{person}{David Marasco}, \bibinfo{person}{Ilya Tyagin}, \bibinfo{person}{Justin Sybrandt}, \bibinfo{person}{James~H Spencer}, {and} \bibinfo{person}{Ilya Safro}.} \bibinfo{year}{2023}\natexlab{}.
\newblock \showarticletitle{Literature-based Discovery for Landscape Planning}.
\newblock \bibinfo{journal}{\emph{arXiv preprint arXiv:2306.02588}} (\bibinfo{year}{2023}).
\newblock


\bibitem[Mikolov et~al\mbox{.}(2013)]%
        {mikolov2013distributed}
\bibfield{author}{\bibinfo{person}{Tomas Mikolov}, \bibinfo{person}{Ilya Sutskever}, \bibinfo{person}{Kai Chen}, \bibinfo{person}{Greg~S Corrado}, {and} \bibinfo{person}{Jeff Dean}.} \bibinfo{year}{2013}\natexlab{}.
\newblock \showarticletitle{Distributed representations of words and phrases and their compositionality}.
\newblock \bibinfo{journal}{\emph{Advances in neural information processing systems}}  \bibinfo{volume}{26} (\bibinfo{year}{2013}).
\newblock


\bibitem[Miranda et~al\mbox{.}(2021)]%
        {miranda2021overview}
\bibfield{author}{\bibinfo{person}{Antonio Miranda}, \bibinfo{person}{Farrokh Mehryary}, \bibinfo{person}{Jouni Luoma}, \bibinfo{person}{Sampo Pyysalo}, \bibinfo{person}{Alfonso Valencia}, {and} \bibinfo{person}{Martin Krallinger}.} \bibinfo{year}{2021}\natexlab{}.
\newblock \showarticletitle{Overview of DrugProt BioCreative VII track: quality evaluation and large scale text mining of drug-gene/protein relations}. In \bibinfo{booktitle}{\emph{Proceedings of the seventh BioCreative challenge evaluation workshop}}. \bibinfo{pages}{11--21}.
\newblock


\bibitem[Piñero et~al\mbox{.}(2019)]%
        {10.1093/nar/gkz1021}
\bibfield{author}{\bibinfo{person}{Janet Piñero}, \bibinfo{person}{Juan~Manuel Ramírez-Anguita}, \bibinfo{person}{Josep Saüch-Pitarch}, \bibinfo{person}{Francesco Ronzano}, \bibinfo{person}{Emilio Centeno}, \bibinfo{person}{Ferran Sanz}, {and} \bibinfo{person}{Laura~I Furlong}.} \bibinfo{year}{2019}\natexlab{}.
\newblock \showarticletitle{{The DisGeNET knowledge platform for disease genomics: 2019 update}}.
\newblock \bibinfo{journal}{\emph{Nucleic Acids Research}} \bibinfo{volume}{48}, \bibinfo{number}{D1} (\bibinfo{date}{11} \bibinfo{year}{2019}), \bibinfo{pages}{D845--D855}.
\newblock
\showISSN{0305-1048}
\urldef\tempurl%
\url{https://doi.org/10.1093/nar/gkz1021}
\showDOI{\tempurl}
\showeprint{https://academic.oup.com/nar/article-pdf/48/D1/D845/31697865/gkz1021.pdf}


\bibitem[Rehurek and Sojka(2011)]%
        {rehurek2011gensim}
\bibfield{author}{\bibinfo{person}{Radim Rehurek} {and} \bibinfo{person}{Petr Sojka}.} \bibinfo{year}{2011}\natexlab{}.
\newblock \showarticletitle{Gensim--python framework for vector space modelling}.
\newblock \bibinfo{journal}{\emph{NLP Centre, Faculty of Informatics, Masaryk University, Brno, Czech Republic}} \bibinfo{volume}{3}, \bibinfo{number}{2} (\bibinfo{year}{2011}).
\newblock


\bibitem[Rindflesch and Fiszman(2003)]%
        {RINDFLESCH2003462}
\bibfield{author}{\bibinfo{person}{Thomas~C Rindflesch} {and} \bibinfo{person}{Marcelo Fiszman}.} \bibinfo{year}{2003}\natexlab{}.
\newblock \showarticletitle{The interaction of domain knowledge and linguistic structure in natural language processing: interpreting hypernymic propositions in biomedical text}.
\newblock \bibinfo{journal}{\emph{Journal of Biomedical Informatics}} \bibinfo{volume}{36}, \bibinfo{number}{6} (\bibinfo{year}{2003}), \bibinfo{pages}{462--477}.
\newblock
\showISSN{1532-0464}
\urldef\tempurl%
\url{https://doi.org/10.1016/j.jbi.2003.11.003}
\showDOI{\tempurl}
\newblock
\shownote{Unified Medical Language System}.


\bibitem[Sedler and Mitchell(2019)]%
        {sedler2019semnet}
\bibfield{author}{\bibinfo{person}{Andrew~R Sedler} {and} \bibinfo{person}{Cassie~S Mitchell}.} \bibinfo{year}{2019}\natexlab{}.
\newblock \showarticletitle{SemNet: Using local features to navigate the biomedical concept graph}.
\newblock \bibinfo{journal}{\emph{Frontiers in Bioengineering and Biotechnology}}  \bibinfo{volume}{7} (\bibinfo{year}{2019}), \bibinfo{pages}{156}.
\newblock


\bibitem[Sollis et~al\mbox{.}(2022)]%
        {10.1093/nar/gkac1010}
\bibfield{author}{\bibinfo{person}{Elliot Sollis}, \bibinfo{person}{Abayomi Mosaku}, \bibinfo{person}{Ala Abid}, \bibinfo{person}{Annalisa Buniello}, \bibinfo{person}{Maria Cerezo}, \bibinfo{person}{Laurent Gil}, \bibinfo{person}{Tudor Groza}, \bibinfo{person}{Osman Güneş}, \bibinfo{person}{Peggy Hall}, \bibinfo{person}{James Hayhurst}, \bibinfo{person}{Arwa Ibrahim}, \bibinfo{person}{Yue Ji}, \bibinfo{person}{Sajo John}, \bibinfo{person}{Elizabeth Lewis}, \bibinfo{person}{Jacqueline A~L MacArthur}, \bibinfo{person}{Aoife McMahon}, \bibinfo{person}{David Osumi-Sutherland}, \bibinfo{person}{Kalliope Panoutsopoulou}, \bibinfo{person}{Zoë Pendlington}, \bibinfo{person}{Santhi Ramachandran}, \bibinfo{person}{Ray Stefancsik}, \bibinfo{person}{Jonathan Stewart}, \bibinfo{person}{Patricia Whetzel}, \bibinfo{person}{Robert Wilson}, \bibinfo{person}{Lucia Hindorff}, \bibinfo{person}{Fiona Cunningham}, \bibinfo{person}{Samuel A Lambert}, \bibinfo{person}{Michael Inouye}, \bibinfo{person}{Helen Parkinson}, {and}
  \bibinfo{person}{Laura W Harris}.} \bibinfo{year}{2022}\natexlab{}.
\newblock \showarticletitle{{The NHGRI-EBI GWAS Catalog: knowledgebase and deposition resource}}.
\newblock \bibinfo{journal}{\emph{Nucleic Acids Research}} \bibinfo{volume}{51}, \bibinfo{number}{D1} (\bibinfo{date}{11} \bibinfo{year}{2022}), \bibinfo{pages}{D977--D985}.
\newblock
\showISSN{0305-1048}
\urldef\tempurl%
\url{https://doi.org/10.1093/nar/gkac1010}
\showDOI{\tempurl}
\showeprint{https://academic.oup.com/nar/article-pdf/51/D1/D977/48440802/gkac1010.pdf}


\bibitem[Sundararajan et~al\mbox{.}(2017)]%
        {sundararajan2017axiomatic}
\bibfield{author}{\bibinfo{person}{Mukund Sundararajan}, \bibinfo{person}{Ankur Taly}, {and} \bibinfo{person}{Qiqi Yan}.} \bibinfo{year}{2017}\natexlab{}.
\newblock \showarticletitle{Axiomatic attribution for deep networks}. In \bibinfo{booktitle}{\emph{International conference on machine learning}}. PMLR, \bibinfo{pages}{3319--3328}.
\newblock


\bibitem[Swanson(1986)]%
        {swanson1986undiscovered}
\bibfield{author}{\bibinfo{person}{Don~R Swanson}.} \bibinfo{year}{1986}\natexlab{}.
\newblock \showarticletitle{Undiscovered public knowledge}.
\newblock \bibinfo{journal}{\emph{The Library Quarterly}} \bibinfo{volume}{56}, \bibinfo{number}{2} (\bibinfo{year}{1986}), \bibinfo{pages}{103--118}.
\newblock


\bibitem[Swanson et~al\mbox{.}(2006)]%
        {swanson2006ranking}
\bibfield{author}{\bibinfo{person}{Don~R Swanson}, \bibinfo{person}{Neil~R Smalheiser}, {and} \bibinfo{person}{Vetle~I Torvik}.} \bibinfo{year}{2006}\natexlab{}.
\newblock \showarticletitle{Ranking indirect connections in literature-based discovery: The role of medical subject headings}.
\newblock \bibinfo{journal}{\emph{Journal of the American society for information science and technology}} \bibinfo{volume}{57}, \bibinfo{number}{11} (\bibinfo{year}{2006}), \bibinfo{pages}{1427--1439}.
\newblock


\bibitem[Sybrandt et~al\mbox{.}(2018a)]%
        {sybrandt2018b}
\bibfield{author}{\bibinfo{person}{Justin Sybrandt}, \bibinfo{person}{Angelo Carrabba}, \bibinfo{person}{Alexander Herzog}, {and} \bibinfo{person}{Ilya Safro}.} \bibinfo{year}{2018}\natexlab{a}.
\newblock \showarticletitle{Are Abstracts Enough for Hypothesis Generation?}. In \bibinfo{booktitle}{\emph{2018 IEEE International Conference on Big Data}}. \bibinfo{pages}{1504--1513}.
\newblock
\urldef\tempurl%
\url{https://doi.org/10.1109/bigdata.2018.8621974}
\showDOI{\tempurl}


\bibitem[Sybrandt and Safro(2021)]%
        {sybrandt2021cbag}
\bibfield{author}{\bibinfo{person}{Justin Sybrandt} {and} \bibinfo{person}{Ilya Safro}.} \bibinfo{year}{2021}\natexlab{}.
\newblock \showarticletitle{CBAG: Conditional biomedical abstract generation}.
\newblock \bibinfo{journal}{\emph{Plos one}} \bibinfo{volume}{16}, \bibinfo{number}{7} (\bibinfo{year}{2021}), \bibinfo{pages}{e0253905}.
\newblock


\bibitem[Sybrandt et~al\mbox{.}(2017)]%
        {sybrandt2017}
\bibfield{author}{\bibinfo{person}{Justin Sybrandt}, \bibinfo{person}{Michael Shtutman}, {and} \bibinfo{person}{Ilya Safro}.} \bibinfo{year}{2017}\natexlab{}.
\newblock \showarticletitle{MOLIERE: Automatic Biomedical Hypothesis Generation System}. In \bibinfo{booktitle}{\emph{Proceedings of the 23rd ACM SIGKDD}} (Halifax, NS, Canada) \emph{(\bibinfo{series}{KDD '17})}. \bibinfo{publisher}{ACM}, \bibinfo{address}{New York, NY, USA}, \bibinfo{pages}{1633--1642}.
\newblock
\showISBNx{978-1-4503-4887-4}
\urldef\tempurl%
\url{https://doi.org/10.1145/3097983.3098057}
\showDOI{\tempurl}


\bibitem[Sybrandt et~al\mbox{.}(2018b)]%
        {sybrandt2018a}
\bibfield{author}{\bibinfo{person}{Justin Sybrandt}, \bibinfo{person}{Micheal Shtutman}, {and} \bibinfo{person}{Ilya Safro}.} \bibinfo{year}{2018}\natexlab{b}.
\newblock \showarticletitle{Large-Scale Validation of Hypothesis Generation Systems via Candidate Ranking}. In \bibinfo{booktitle}{\emph{2018 IEEE International Conference on Big Data}}. \bibinfo{pages}{1494--1503}.
\newblock
\urldef\tempurl%
\url{https://doi.org/10.1109/bigdata.2018.8622637}
\showDOI{\tempurl}


\bibitem[Sybrandt et~al\mbox{.}(2020)]%
        {sybrandt2020agatha}
\bibfield{author}{\bibinfo{person}{Justin Sybrandt}, \bibinfo{person}{Ilya Tyagin}, \bibinfo{person}{Michael Shtutman}, {and} \bibinfo{person}{Ilya Safro}.} \bibinfo{year}{2020}\natexlab{}.
\newblock \showarticletitle{AGATHA: Automatic Graph Mining And Transformer Based Hypothesis Generation Approach}. In \bibinfo{booktitle}{\emph{Proceedings of the 29th ACM International Conference on Information \& Knowledge Management}}. \bibinfo{pages}{2757--2764}.
\newblock


\bibitem[Szklarczyk et~al\mbox{.}(2020)]%
        {10.1093/nar/gkaa1074}
\bibfield{author}{\bibinfo{person}{Damian Szklarczyk}, \bibinfo{person}{Annika~L Gable}, \bibinfo{person}{Katerina~C Nastou}, \bibinfo{person}{David Lyon}, \bibinfo{person}{Rebecca Kirsch}, \bibinfo{person}{Sampo Pyysalo}, \bibinfo{person}{Nadezhda~T Doncheva}, \bibinfo{person}{Marc Legeay}, \bibinfo{person}{Tao Fang}, \bibinfo{person}{Peer Bork}, \bibinfo{person}{Lars~J Jensen}, {and} \bibinfo{person}{Christian von Mering}.} \bibinfo{year}{2020}\natexlab{}.
\newblock \showarticletitle{{The STRING database in 2021: customizable protein–protein networks, and functional characterization of user-uploaded gene/measurement sets}}.
\newblock \bibinfo{journal}{\emph{Nucleic Acids Research}} \bibinfo{volume}{49}, \bibinfo{number}{D1} (\bibinfo{date}{11} \bibinfo{year}{2020}), \bibinfo{pages}{D605--D612}.
\newblock
\showISSN{0305-1048}
\urldef\tempurl%
\url{https://doi.org/10.1093/nar/gkaa1074}
\showDOI{\tempurl}
\showeprint{https://academic.oup.com/nar/article-pdf/49/D1/D605/40395991/gkaa1074.pdf}


\bibitem[Tyagin et~al\mbox{.}(2022)]%
        {tyagin2022accelerating}
\bibfield{author}{\bibinfo{person}{Ilya Tyagin}, \bibinfo{person}{Ankit Kulshrestha}, \bibinfo{person}{Justin Sybrandt}, \bibinfo{person}{Krish Matta}, \bibinfo{person}{Michael Shtutman}, {and} \bibinfo{person}{Ilya Safro}.} \bibinfo{year}{2022}\natexlab{}.
\newblock \showarticletitle{Accelerating COVID-19 research with graph mining and transformer-based learning}. In \bibinfo{booktitle}{\emph{Proceedings of the AAAI Conference on Artificial Intelligence}}, Vol.~\bibinfo{volume}{36}. \bibinfo{pages}{12673--12679}.
\newblock


\bibitem[Tyagin and Safro(2021)]%
        {tyagin2021interpretable}
\bibfield{author}{\bibinfo{person}{Ilya Tyagin} {and} \bibinfo{person}{Ilya Safro}.} \bibinfo{year}{2021}\natexlab{}.
\newblock \showarticletitle{Interpretable visualization of scientific hypotheses in literature-based discovery}.
\newblock \bibinfo{journal}{\emph{BioCretive Workshop VII}} (\bibinfo{year}{2021}).
\newblock


\bibitem[Ursu et~al\mbox{.}(2016)]%
        {10.1093/nar/gkw993}
\bibfield{author}{\bibinfo{person}{Oleg Ursu}, \bibinfo{person}{Jayme Holmes}, \bibinfo{person}{Jeffrey Knockel}, \bibinfo{person}{Cristian~G. Bologa}, \bibinfo{person}{Jeremy~J. Yang}, \bibinfo{person}{Stephen~L. Mathias}, \bibinfo{person}{Stuart~J. Nelson}, {and} \bibinfo{person}{Tudor~I. Oprea}.} \bibinfo{year}{2016}\natexlab{}.
\newblock \showarticletitle{{DrugCentral: online drug compendium}}.
\newblock \bibinfo{journal}{\emph{Nucleic Acids Research}} \bibinfo{volume}{45}, \bibinfo{number}{D1} (\bibinfo{date}{10} \bibinfo{year}{2016}), \bibinfo{pages}{D932--D939}.
\newblock
\showISSN{0305-1048}
\urldef\tempurl%
\url{https://doi.org/10.1093/nar/gkw993}
\showDOI{\tempurl}
\showeprint{https://academic.oup.com/nar/article-pdf/45/D1/D932/8847374/gkw993.pdf}


\bibitem[Welling and Kipf(2016)]%
        {welling2016semi}
\bibfield{author}{\bibinfo{person}{Max Welling} {and} \bibinfo{person}{Thomas~N Kipf}.} \bibinfo{year}{2016}\natexlab{}.
\newblock \showarticletitle{Semi-supervised classification with graph convolutional networks}. In \bibinfo{booktitle}{\emph{J. International Conference on Learning Representations (ICLR 2017)}}.
\newblock


\bibitem[Xing et~al\mbox{.}(2020)]%
        {xing2020biorel}
\bibfield{author}{\bibinfo{person}{Rui Xing}, \bibinfo{person}{Jie Luo}, {and} \bibinfo{person}{Tengwei Song}.} \bibinfo{year}{2020}\natexlab{}.
\newblock \showarticletitle{BioRel: towards large-scale biomedical relation extraction}.
\newblock \bibinfo{journal}{\emph{BMC bioinformatics}} \bibinfo{volume}{21}, \bibinfo{number}{16} (\bibinfo{year}{2020}), \bibinfo{pages}{1--13}.
\newblock


\bibitem[Xun et~al\mbox{.}(2017)]%
        {8215526}
\bibfield{author}{\bibinfo{person}{Guangxu Xun}, \bibinfo{person}{Kishlay Jha}, \bibinfo{person}{Vishrawas Gopalakrishnan}, \bibinfo{person}{Yaliang Li}, {and} \bibinfo{person}{Aidong Zhang}.} \bibinfo{year}{2017}\natexlab{}.
\newblock \showarticletitle{Generating Medical Hypotheses Based on Evolutionary Medical Concepts}. In \bibinfo{booktitle}{\emph{2017 IEEE International Conference on Data Mining (ICDM)}}. \bibinfo{pages}{535--544}.
\newblock
\urldef\tempurl%
\url{https://doi.org/10.1109/ICDM.2017.63}
\showDOI{\tempurl}


\bibitem[Zeng et~al\mbox{.}(2006)]%
        {zeng2006rxnav}
\bibfield{author}{\bibinfo{person}{Kelly Zeng}, \bibinfo{person}{Olivier Bodenreider}, \bibinfo{person}{John Kilbourne}, {and} \bibinfo{person}{Stuart~J Nelson}.} \bibinfo{year}{2006}\natexlab{}.
\newblock \showarticletitle{RxNav: a web service for standard drug information}. In \bibinfo{booktitle}{\emph{AMIA Annual Symposium Proceedings}}, Vol.~\bibinfo{volume}{2006}. American Medical Informatics Association, \bibinfo{pages}{1156}.
\newblock


\end{thebibliography}

%%\appendix

\ifthesis
\else
\pagebreak

\fi

\end{document}